\documentclass{article}
\PassOptionsToPackage{authoryear,round}{natbib}

\usepackage[preprint]{neurips_2026}

\usepackage[utf8]{inputenc} 
\usepackage[T1]{fontenc}    
\usepackage{hyperref}       
\usepackage{url}            
\usepackage{booktabs}       
\usepackage{amsfonts}       
\usepackage{nicefrac}       
\usepackage{microtype}      
\usepackage{xcolor}         
\usepackage{soul}
\usepackage{booktabs}
\usepackage{lineno}
\usepackage{subfigure}
\usepackage{soul}
\usepackage{booktabs}
\usepackage{amsmath}
\usepackage{graphicx}
\usepackage{cleveref}
\usepackage{bm}

\title{Single-Shot HDR Recovery via a Video Diffusion Prior }

\DeclareMathOperator*{\argmin}{arg\,min}
\definecolor{myorange}{HTML}{ff9900}

\author{%
  Chinmay Talegaonkar\textsuperscript{1}\thanks{Corresponding author: \texttt{ctalegaonkar@ucsd.edu}}
  \quad Jinshi He\textsuperscript{1}
  \quad Christopher McKenna\textsuperscript{2}
  \quad Nicholas Antipa\textsuperscript{1} \\
  \textsuperscript{1}University of California San Diego \\
  \textsuperscript{2}Creare LLC
}
\begin{document}
\maketitle
\begin{abstract}
Recent generative methods for single-shot high dynamic range (HDR) image reconstruction show promising results, but often struggle with preserving fidelity to the input image. They require separate models to handle highlights and shadows, or sacrifice interpretability by directly predicting the final HDR image. We address these limitations by re-casting single-shot HDR reconstruction as conditional video generation and fusing the generated frames into an HDR image. We fine-tune a video diffusion model to generate an exposure bracket, conditioned on a low dynamic range (LDR) input. We fuse this image bracket using per-pixel weights predicted by a light-weight UNet. This formulation is simple, interpretable, and effective. Rather than directly hallucinating an HDR image, it explicitly reconstructs the intermediate exposure stack and fuses it into the final output. Our method eliminates the need for separate models across exposure regimes and produces HDR reconstructions with high input fidelity. On quantitative benchmarks, we outperform state-of-the-art generative baselines with comparable model capacity on several reconstruction metrics. Human evaluators further prefer our results in $72\%$ of pairwise comparisons against existing methods. Finally, we show that this input-conditioned sequence generation and fusion framework extends beyond HDR to other image reconstruction tasks, such as all-in-focus image recovery from a single defocus-blurred input.
\end{abstract}
\section{Introduction}
\vspace{-1mm}
High dynamic range (HDR) imaging mitigates the gap between digital imagery and human perception of the real world~\citet{mccann2011art}. It powers richer visual experiences in photography and displays, and supports high-fidelity imaging in medical and industrial settings~\citet{Cao:18}.

\begin{figure}
    \centering
    \includegraphics[width=\linewidth]{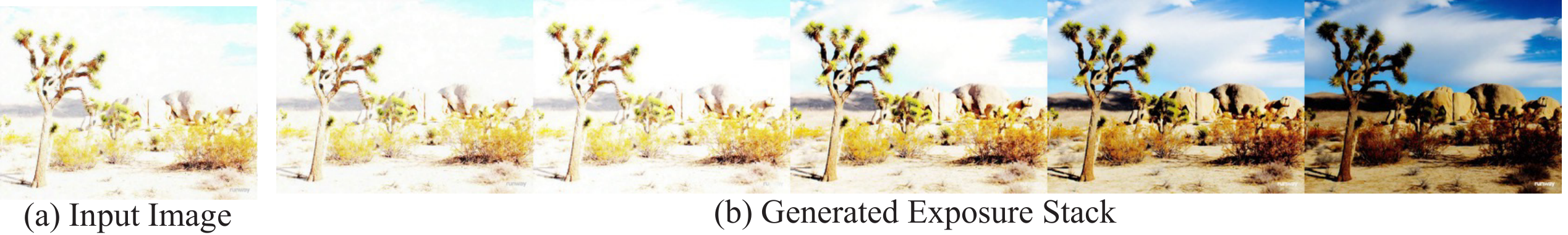}
    \caption{\textbf{Video diffusion models contain implicit priors to generate exposure bursts.} Given an input LDR image (a) captured at a high exposure, we prompt an off-the-shelf video model (Runway Gen 4): "Reduce the brightness of the image over the course of the video, generating an exposure burst stack from high to low exposure". The resulting video frames are plotted in (b), demonstrating that the video model contains strong lighting priors to simulate an exposure bracket. }
    \label{fig:fig1_model_prior}
\end{figure}

Capturing an HDR image usually involves acquiring multiple low dynamic range (LDR) images at different exposures and fusing them into a single HDR output~\citet{mantiuk2015high, debevec1997recovering, kalantari2017deep, wu2018deep, jiang2023meflut, kong2024safnet, yan2019attention}. While effective, this method introduces practical challenges due to motion during capture. Single-shot HDR methods recover an HDR image from a single LDR capture, eliminating temporal artifacts. However, single-shot HDR is fundamentally ill-posed, as it requires hallucinating missing information in the over/under-exposed regions of the input LDR measurement.
We propose a two-stage pipeline for single-shot HDR reconstruction by reframing it as a conditional video generation problem. Our central hypothesis is that an exposure bracket is structurally analogous to a video where instead of scene motion, the temporal variation reflects changes in exposure. We observe that an off-the-shelf image-to-video generation model, with an appropriate prompt, produces coherent bracketed exposures (\cref{fig:fig1_model_prior}). This suggests that video diffusion models~\citet{wan2025wan, blattmann2023stable} can serve as strong priors for synthesizing plausible exposure stacks from a single image. Because video models allow inter-frame motion, which is undesirable in HDR imaging, we fine-tune a video diffusion model~\citet{blattmann2023stable} to generate a motion-free LDR bracket. Unlike recent generative single-shot HDR methods~\citet{wang2025lediff, bemana2025bracket, wu2026x2hdr}, which rely only on spatial priors, our approach leverages both spatial and temporal priors by using a video generation model to generate a plausible image sequence conditioned on the input LDR image. We then fuse the synthesized LDR bracket using per-pixel blend weights predicted by a small network. 

On the standard SI-HDR benchmark~\citet{hanji2022comparison}, our method outperforms generative baselines with comparable model capacity on several reconstruction metrics and remains competitive on the remaining metrics. Human evaluators further prefer our results in $72\%$ of pairwise comparisons against baseline methods. Our framework of input-conditioned image sequence generation followed by light-weight fusion naturally extends to other image recovery tasks. We demonstrate this with all-in-focus recovery from a single defocus-blurred image.
\section{Related Work}
\vspace{-1mm}
\label{sec:related}
\paragraph{Multi-shot HDR Imaging:}
Classical HDR methods capture successive LDR images (exposure bracket) of a scene at different exposure settings. Debevec Merging~\citet{debevec1997recovering} recovers an HDR radiance map from an exposure bracket with known relative exposure times. Mertens Fusion~\citet{mertens2007exposure} blends the exposure stack using per-pixel weights based on contrast, well-exposedness, and saturation. These methods require multiple aligned captures, making them less suitable for dynamic scenes and handheld capture.
Subsequent learning-based methods address motion artifacts using learned alignment, attention, or fusion networks~\citet{kalantari2017deep, wu2018deep, yan2019attention, niu2021hdr}. However, they still require capturing an exposure burst. In contrast, we infer the exposure bracket from a single LDR image.

\paragraph{Single-Image HDR Reconstruction:}
Single-shot HDR methods predict HDR from a single LDR capture. 
HDRCNN~\citet{eilertsen2017hdr} and ExpandNet~\citet{marnerides2018expandnet} trained convolutional networks to hallucinate missing information in over-exposed regions. SingleHDR~\citet{liu2020single} decomposed the camera imaging pipeline and learned to invert each stage, while~\citet{le2023single} proposed generating multiple virtual exposures from a single input in a weakly supervised manner. Several works~\citet{dai2025single,nayar2000high,metzler2020deep,sun2020learning,shi2024split,so2022mantissacam} have explored hardware modifications for single-shot HDR capture, using specialized optics and sensors. In contrast, our method can be applied to raw images captured by standard cameras.

More recently, diffusion models have shown strong potential for single-image HDR generation. LEDiff~\citet{wang2025lediff} uses latent diffusion to separately generate highlight and shadow exposures, which are then fused into an HDR image. UltraFusion~\citet{chen2025ultrafusion} formulates exposure fusion as a guided inpainting problem using a diffusion prior, enabling fusion across large exposure gaps. X2HDR~\citet{wu2026x2hdr} adapts pre-trained diffusion models to HDR generation by operating in a perceptually uniform color space, while UltraLED~\citet{meng2025ultraled} works in the raw domain for ultra-high dynamic range recovery. Notably, these generative approaches rely on \emph{image} diffusion models. In contrast, our work leverages \emph{video} diffusion models. This is motivated by the observation that HDR recovery can be viewed as generating an ordered sequence of images with monotonically increasing exposures, which aligns with the temporal structure modeled by video priors.
\paragraph{Video Diffusion Models for Computational Photography:} 
Recent work shows that video generation models can be repurposed for computational photography. VDM-EVFI~\citet{chen2025repurposing} adapts video diffusion priors for event-based video frame interpolation. Blur2Vid~\citet{tedla2025generating} and Learning to Refocus~\citet{tedla2025learning} fine-tune video models to recover sharp videos from a single motion-blurred image and to refocus images, respectively. These works suggest that video diffusion models encode useful priors over visual attributes that vary across frames. We build on this idea by modeling an HDR exposure stack as a video-like sequence with monotonically increasing exposure. Concurrent works~\citet{yu2026diffhdr,korem2026hdr} fine-tune video models to generate HDR videos from LDR videos. In contrast, we use video diffusion for single-image HDR reconstruction by emulating the multi-exposure capture process.
\section{Preliminaries}
\vspace{-1mm}
\label{sec:prelim}
\paragraph{HDR Imaging:}
Camera sensors capture only a limited portion of the full dynamic range of a scene.  Consequently, images acquired at very low or very high exposure times exhibit underexposure (lose shadow details) or oversaturation artifacts (lose highlight details), respectively. Formally, an image ($I$) captured with an exposure time $t$ is related to the underlying scene radiance $I_{HDR}$ as 
\begin{equation}
    I = g(\textrm{clamp}(t\cdot I_{HDR} + \eta, I_{-}, I_{+})),
\end{equation}
where $I_{-}$ and $I_{+}$ denote the lower and upper saturation thresholds of the sensor, respectively. The function $g(\cdot)$ denotes a combination of the camera response function (CRF) and sensor quantization, and $\eta$ approximates the sensor read noise and photon shot noise. As a result, different regions of the captured image may become under-exposed or over-exposed depending on the chosen exposure time. High exposures can lead to saturation, while low exposures produce noisy quantized measurements. We approximate the CRF as a gamma mapping \(g(x)=x^{1/\gamma}\), where \(\gamma\) denotes the gamma parameter of the linear-to-sRGB transformation \citet{chang2025gcc}.

In addition to capture-time limitations, the limited dynamic range of conventional displays requires HDR images to be \textit{tone-mapped} for visualization~\citet{reinhard2002parameter}. Tone-mapping applies a contraction mapping to the HDR image to reduce its dynamic range to match the display.

Recovering the full dynamic range of a scene, therefore, typically requires capturing and fusing an exposure bracket $\{I_{t_1}, I_{t_2}, I_{t_3}, \dots, I_{t_n}\}$, to reconstruct the HDR image $I_{HDR}$: 
\begin{equation}
    I_{HDR} = f(I_{t_1}, I_{t_2}, I_{t_3}\dots I_{t_n})
\end{equation}
\noindent The fusion function $f(\cdot)$ can be a classical \citet{mertens2007exposure, debevec1997recovering}  or a learning-based approach \citet{kalantari2017deep,yan2019attention, niu2021hdr}.
Single-shot HDR methods are an extreme version of this problem, where $n=1$. 

\paragraph{Latent Video Diffusion Models:}
Latent video diffusion models~\citet{blattmann2023stable, wan2025wan} consist of an encoder $\mathcal{E}$ and a decoder $\mathcal{D}$, which map an RGB video $\mathbf{x}$ to and from a lower-dimensional latent space, respectively. A diffusion model $\phi(\cdot)$ is then trained to operate in this latent space. Given an input condition $c$, such as a text prompt or an image, inference starts with a randomly sampled noise vector $\mathbf{z}_{K} \sim \mathcal{N}(0, I)$. The diffusion model $\phi(\cdot)$ then iteratively denoises $\mathbf{z}_K$ over $K$ steps to produce a clean latent $\mathbf{z}_0$. The clean latent is decoded into an RGB video as $\mathcal{D}(\mathbf{z}_0)$. In this work, we use Stable Video Diffusion~\citet{blattmann2023stable}, which is an image-conditioned latent video diffusion model.
Training follows the EDM framework~\citet{karras2022elucidating}. Given a clean latent $\mathbf{z}=\mathcal{E}(\mathbf{x})$, Gaussian noise is added at a denoising-step-dependent noise level $\sigma_k$ to obtain
\begin{equation}
\mathbf{z}_{k} = \mathbf{z} + \sigma_k \boldsymbol{\epsilon}, 
\qquad \boldsymbol{\epsilon} \sim \mathcal{N}(0,I).
\label{eq:noisy_latent}
\end{equation}
\noindent The denoised latent is predicted using the EDM parameterization
\begin{equation}
\hat{\mathbf{z}} = c_{\mathrm{skip}}(\sigma_k)\,\mathbf{z}_{k}
+ c_{\mathrm{out}}(\sigma_k)\, f_\theta(\mathbf{z}_{k}, \sigma_k, c),
\label{eq:edm_denoised}
\end{equation}
where $f_\theta$ denotes the spatio-temporal denoiser and $c$ the conditioning signal. The coefficients $c_{\mathrm{skip}}(\sigma_k)$ and $c_{\mathrm{out}}(\sigma_k)$ are noise-dependent preconditioning terms that modulate the contributions of the noisy latent and the denoiser output. The training process minimizes
\begin{equation}
\mathcal{L} = \mathbb{E}_{\mathbf{z},\boldsymbol{\epsilon},\sigma_k}
\left[
w(\sigma_k)\,\|\hat{\mathbf{z}}-\mathbf{z}\|_2^2
\right],
\label{eq:edm_loss}
\end{equation}
with $w(\sigma_k)$ being a noise-dependent weighting function.

\section{Method} \label{sec:method}
\vspace{-1mm}
\begin{figure}
    \centering
    \includegraphics[width=1.0\linewidth]{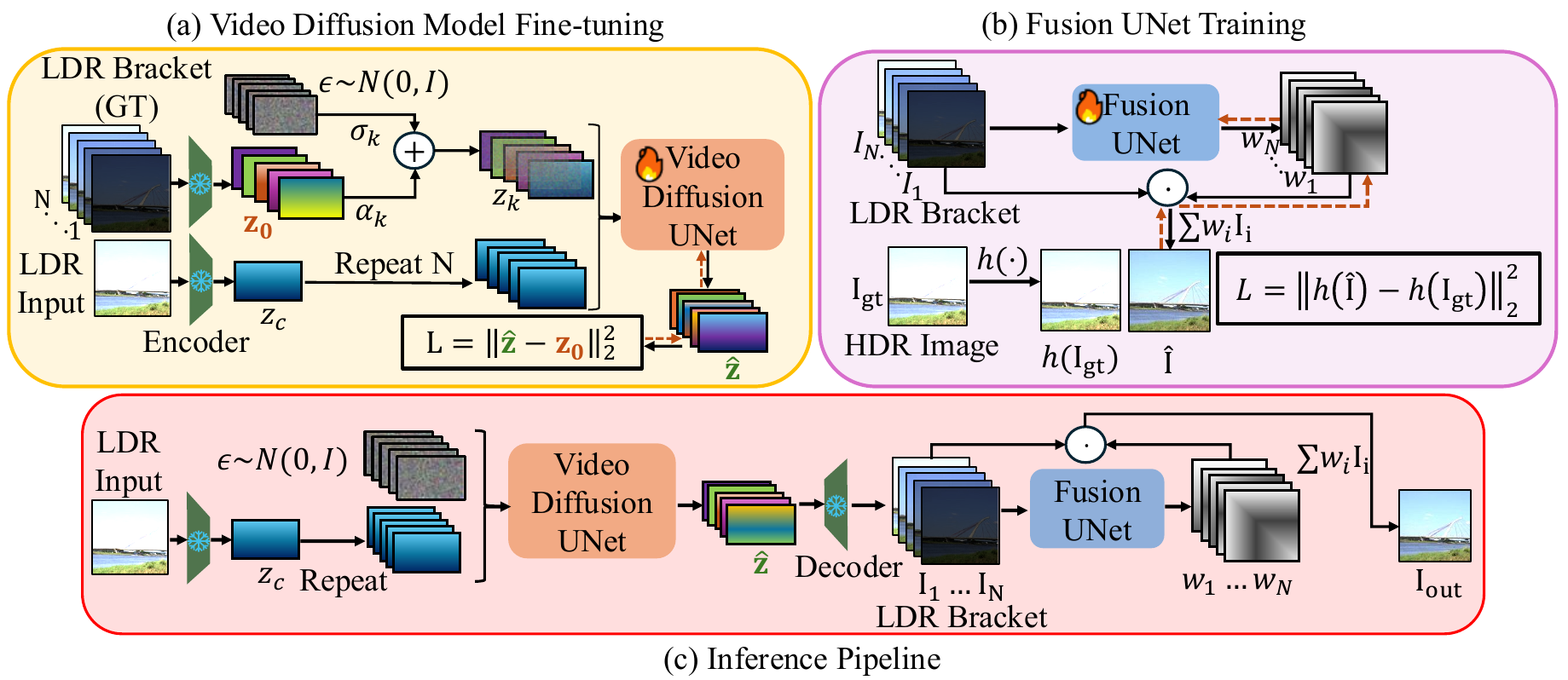}
    \caption{\textbf{Method overview.} 
(a) Given an input LDR image, we fine-tune a latent video diffusion model to generate an LDR exposure bracket corresponding to the scene. (b) A fusion UNet then predicts per-pixel weights to fuse this bracket into an HDR image. The fusion network is supervised in PU-21 encoded space, and at (c) inference time, we sample the video diffusion model to synthesize an LDR bracket, which is fused by the Fusion UNet to obtain the final HDR output.}
    \label{fig:method}
\end{figure}
We recast single-shot HDR reconstruction as conditional exposure-bracket generation from a single LDR image. Since adjacent exposures form a coherent video-like sequence, we fine-tune a video diffusion model to synthesize the full exposure bracket. During training, each HDR image is converted into an LDR exposure bracket comprising monotonically increasing exposures. As a result, the model always learns a canonical mapping from low to high exposures along the temporal axis, even when the observed conditioning frame is not the first frame in the sequence. We then train a separate, smaller network (Fusion UNet) to predict per-pixel weights for fusing the LDR frames generated by the fine-tuned video model. Below, we outline our two-stage training process (\cref{fig:method}), which consists of fine-tuning the video model and training the Fusion UNet. 

\paragraph{Video model fine-tuning algorithm:} Given a single input LDR image at an arbitrary exposure, we fine-tune the video diffusion model to generate a plausible sequence of LDR images with monotonically increasing exposures that covers the scene’s dynamic range. Training data consists of pairs of raw images and their corresponding ground truth (GT) HDR images, sourced from \citet{fairchild2007hdr, zou2023rawhdr}, following~\citet{wu2026x2hdr}. 
Given a GT HDR image $I_{HDR}$, the corresponding LDR image captured at an exposure $t$ can be obtained as \begin{equation}
    I_t = (\textrm{clamp}(t\cdot I_{HDR}, 0, 1))^{1/\gamma}
    \label{eq:HDR2LDR}
\end{equation}

We use $\gamma=2.2$ as a simple approximation to the nonlinear tone response of sRGB-like images, which dominate the internet-scale pretraining data of the base video model. We generate the GT LDR bracket over an exposure range $[t_s, t_e]$ determined from the GT HDR image. To compute these exposure bounds, we first estimate the luminance map
\[
Y = 0.2126\, I^r_{HDR} + 0.7152\, I^g_{HDR} + 0.0722\, I^b_{HDR},
\]
where $I^{\{r,g,b\}}_{HDR}$ denote the red, green, and blue channels of $I_{HDR}$. Following prior work~\citet{wang2025lediff}, we set
\begin{equation}
t_s = \frac{v_s^{\gamma}}{Y_{\max}}, \qquad t_e = \frac{v_e^{\gamma}}{Y_{\text{median}}},
\label{eq:expselection}
\end{equation}
\noindent where $Y_{\max}$ and $Y_{\text{median}}$ are the maximum and median values of $Y$. The target LDR intensities $v_s,v_e \in (0,1]$ control the dynamic range mapping, and are both set to $0.85$ in all experiments. This choice maps the brightest pixel and the median-luminance pixel to $v_s$ and $v_e$, respectively, after applying the $\gamma$ curve. We then sample $N=5$ exposure times uniformly in log-space between $\log_2 t_s$ and $\log_2 t_e$, and generate the corresponding noiseless LDR bracket using \cref{eq:HDR2LDR}.

To fine-tune the video model, we first encode the raw LDR input image \(I_c\) into a latent vector \(\mathbf{z}_c = \mathcal{E}(I_c)\), and encode the simulated GT LDR exposure bracket into latent vectors \(\mathbf{z}_1, \mathbf{z}_2, \dots, \mathbf{z}_N\).
Importantly, the input \(\mathbf{z}_k\) to the diffusion model is obtained by noising only the \emph{clean} exposure-bracket latents, while the model is conditioned on the latent \(\mathbf{z}_c\) corresponding to the input raw image $I_c$. We do not use classifier-free guidance~\citet{ho2022classifier} (CFG) during inference and accordingly disable conditional dropout during training. This follows prior work on repurposing diffusion models for downstream vision tasks~\citet{ke2024repurposing}. Unlike text-conditioned generation, where CFG is often used to improve sample quality and diversity, our task requires strong pixel-level consistency with the input condition. In this setting, we find that omitting CFG leads to better reconstruction performance.

For fine-tuning the video diffusion UNet, we concatenate \(\mathbf{z}_c\) and the noisy latent \(\mathbf{z}_k\) along the channel dimension and feed the resulting tensor to the model, similar to prior conditioning schemes used in related work~\citet{chen2025repurposing, tedla2025learning}. The diffusion UNet is then trained to minimize \cref{eq:edm_loss}. Compared to prior HDR reconstruction methods that either rely on separate models for shadow and highlight recovery~\citet{wang2025lediff} or directly predict HDR-space representations from a single image~\citet{wu2026x2hdr}, our approach retains simplicity and interpretability by using a single model to generate the full LDR exposure bracket, which is then fused into the final HDR image using per-pixel weights.

\paragraph{Fusion UNet:}
We train a separate fusion network \(U_{\theta}\) that takes as input an LDR exposure bracket \(\mathbf{I} = [I_1, I_2, \dots, I_N]\) and predicts a corresponding stack of per-pixel fusion weights \(\mathbf{W} = [W_1, W_2, \dots, W_N]\), such that  $\sum_{i=1}^{N} W_i(\mathbf{p}) = 1
\quad \forall \mathbf{p}$, where \(\mathbf{p}\) indexes pixel location. The fusion network is trained independently using LDR brackets synthesized from clean ground-truth HDR images \(I_{\mathrm{HDR}}\) via the HDR-to-LDR mapping in \cref{eq:HDR2LDR} and the exposure sampling strategy in \cref{eq:expselection}. As a result, it is trained on the same target distribution of LDR brackets as the video diffusion model. 
The fused prediction is then given by
\begin{equation}
    \hat{I} = \sum_{i=1}^{N} W_i \odot I_i,
\end{equation}
where \(\odot\) denotes element-wise multiplication. The input LDR frames are linearized during both training and inference using the inverse gamma mapping, \(x\rightarrow x^\gamma\), so that the fused output is a linear radiance image. We scale the ground-truth HDR $I_{HDR}$ by the inverse of its peak luminance, resulting in a normalized $\tilde{I}_{HDR}$. As a result, our model predicts HDR only up to scale, following prior work~\citet{wang2025lediff,wu2026x2hdr}. We train the fusion network by minimizing
\begin{equation}
\mathcal{L}_{\mathrm{fusion}}(\theta)
=
\left\|
h(\hat{I}) - h(\tilde{I}_{\mathrm{HDR}})
\right\|_2^2,
\end{equation}
where \(h(\cdot)\) denotes the PU-21 transformation~\citet{wu2026x2hdr,ke2025training} 
\begin{equation}
h(L)=a\bigl(\log_2 L - L_{\min}\bigr)^2 + b\bigl(\log_2 L - L_{\min}\bigr),
\end{equation}
which maps HDR intensities to a normalized perceptual space with greater precision in low-luminance regions. We use the same hyperparameter values \((a,b,L_{\min})\) as prior work~\citet{wu2026x2hdr}. Other log-based HDR transformations~\citet{kalantari2017deep,yu2026diffhdr} could also be used in place of PU-21. This loss encourages the fusion network to predict weights that produce perceptually and quantitatively accurate  HDR reconstructions up to a scaling factor. 
\paragraph{Why a light-weight fusion network?} We intentionally choose a light-weight fusion network that predicts only per-pixel fusion weights, restricting fusion to a linear aggregation of the generated exposure bracket. Strong HDR reconstruction under this constrained design provides indirect evidence that the synthesized brackets contain useful HDR information. Please see \autoref{app:K} for additional analysis and visualizations of the synthesized exposure brackets by the fine-tuned video model.

\paragraph{Implementation Details:}
All models are trained at an image resolution of \(512 \times 512\). We use Stable Video Diffusion (SVD) as the video diffusion backbone with \(N=5\) frames per sequence. We fine-tune the diffusion UNet (\(1.5\)B parameters) of SVD for 16K iterations using AdamW with a learning rate of \(5\times10^{-5}\) and 100 warm-up steps. Training uses full FP32 precision with gradient clipping at 1.0. We use 8 AMD MI300X GPUs with a batch size of 1 per GPU and gradient accumulation over 8 steps. The Fusion UNet is a small UNet-based model~\citet{ronneberger2015u} with \(\sim 0.5\)M parameters, trained for 25K iterations on 4 MI250 GPUs with total batch size 64 and learning rate \(1\times10^{-5}\). Total training time is approximately 36 hours for the video diffusion UNet and 12 hours for the Fusion UNet. During inference, our method requires approximately 13.5 GB of GPU VRAM. 
\section{Experiments and Results}
\vspace{-1mm}
\label{sec:experiments}
\paragraph{Training and evaluation data:} We generate training data by combining HDR images from the same sources as X2HDR~\citet{wu2026x2hdr}, resulting in a total of approximately 11K training crops. Please see \autoref{app:A} for more details on the crop extraction process. We use the SI-HDR dataset~\citet{hanji2022comparison} for evaluation, adopting the same subset as X2HDR~\citet{wu2026x2hdr}. The scenes cover diverse indoor, outdoor, natural, and urban content. Notably, some of the raw images contain line noise artifacts, which pose an additional challenge for HDR reconstruction.
\noindent \paragraph{Quantitative HDR evaluation:}
We follow the evaluation criteria~\citet{cao2024perceptual} ($Q^{\ast}$ metrics) used in prior work~\citet{wu2026x2hdr}. The $Q^{\ast}$ metrics circumvent the global exposure difference between the ground truth and the predicted HDR image by optimizing exposure values that maximize performance for a given metric. We evaluate $Q^{\ast}$ metrics for PSNR, SSIM, LPIPS, and MAE. Additionally, we compute the $Q$-JOD metric from HDR-VDP3~\citet{mantiuk2023hdr}, which is known to correlate better with human preference. 
See \autoref{app:B} for more evaluation details. We compare our method with several kinds of baselines: diffusion-based generative methods with comparable model capacity, such as X2HDR~\citet{wu2026x2hdr} and LEDiff~\citet{wang2025lediff}; BracketDiff~\citet{bemana2025bracket}, which is a diffusion-based inference-time optimization method; and learning-based feedforward methods like SingleHDR~\citet{liu2020single} and HDRCNN~\citet{eilertsen2017hdr}.

\begin{figure}[t]
    \centering
    \includegraphics[width=0.95\linewidth]{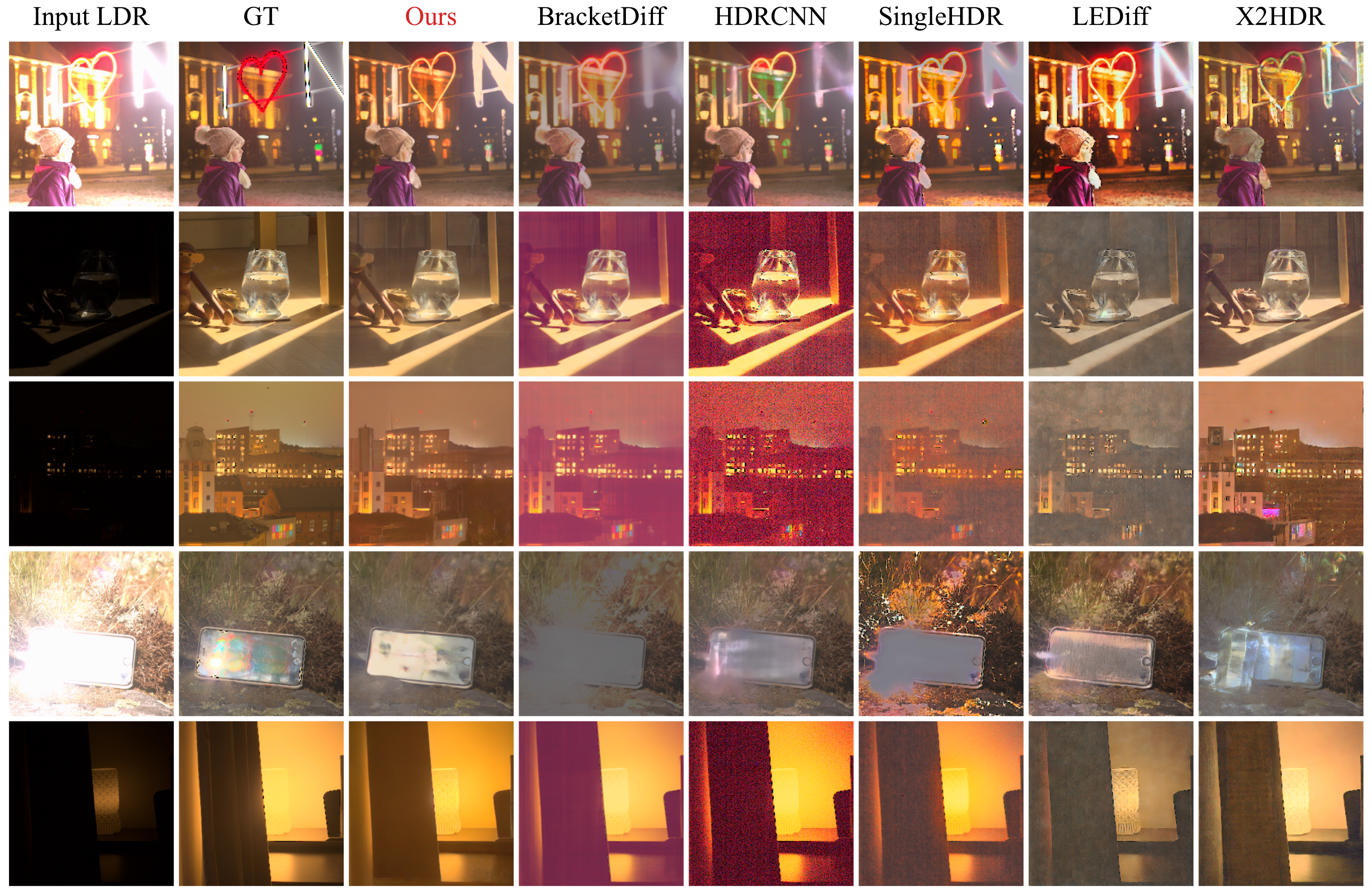}
    \caption{\textbf{Qualitative HDR reconstruction results.} We show challenging ill-exposed examples where our method produces higher-fidelity HDR reconstructions than the baselines. Our method is better at denoising under-exposed regions (rows 2, 3, 4) and generates more plausible content in over-exposed regions (rows 1, 4). All results are normalized to the same median luminance and visualized with Reinhard tone mapping~\citet{reinhard2002parameter}.}
    \label{fig:main-results}
\end{figure}

\begin{figure}[t]
    \centering
    \begin{minipage}{0.58\linewidth}
        \centering
        \includegraphics[width=\linewidth]{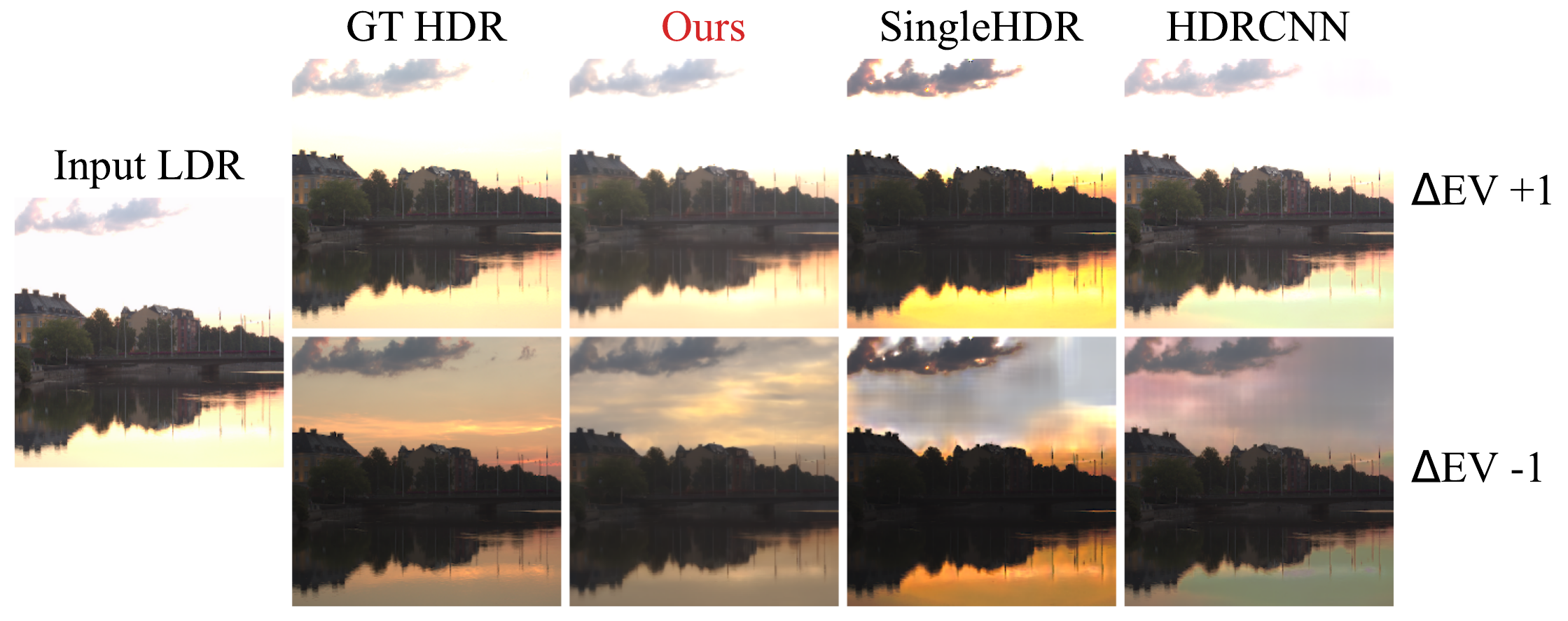}
        \caption{\textbf{Non-generative methods struggle with over-exposed regions.} Our generative prior produces plausible details for over-exposed regions as visible at a lower exposure image (row 2), while SingleHDR and HDRCNN produce artifacts (zoom in on clouds).}
        \label{fig:hallucination-ev}
    \end{minipage}
    \hfill
    \begin{minipage}{0.38\linewidth}
        \centering
        \includegraphics[width=\linewidth]{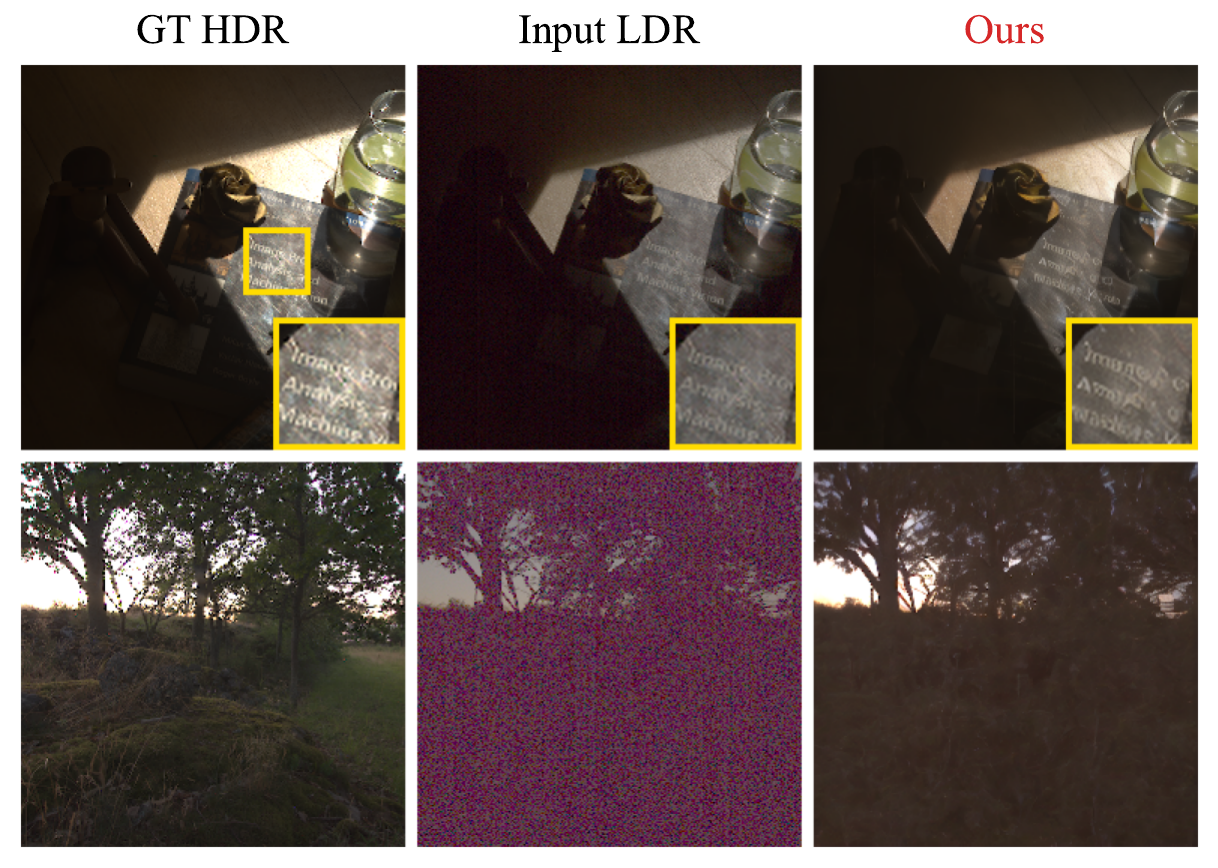}
        \caption{\textbf{Failure cases}. Our method can struggle with fine text (row 1), as the VAE encoding can lose out on small, high-frequency details. Extremely dark inputs (row 2) remain challenging for all methods due to a low signal-to-noise ratio.}
        \label{fig:failure-cases}
    \end{minipage}
\end{figure}

We show in \cref{tab:hdr-comparison} that our method achieves strong performance against state-of-the-art generative baselines, outperforming them on several reconstruction metrics ($Q^{\ast}$-PSNR, $Q^{\ast}$-MAE, and $Q^{\ast}$-LPIPS). While BracketDiff achieves the best $Q^{\ast}$-SSIM, it is $25\times$ slower than our method. Non-generative feedforward methods such as HDRCNN and SingleHDR are faster and perform marginally better on some metrics (Q-JOD, $Q^{\ast}$-SSIM), but are less robust in challenging exposure regimes. In under-exposed scenes, they often fail to suppress raw sensor noise (see \cref{fig:main-results}), while in over-exposed regions they can hallucinate unrealistic structure (see \cref{fig:hallucination-ev}). By training directly on raw inputs, our method better handles noisy under-exposed images while maintaining scene consistency across the synthesized exposure bracket. Although X2HDR is also trained on similar raw-paired data, our method better preserves plausible scene structure in saturated regions (row 4, \cref{fig:main-results}) and achieves stronger aggregate performance. Overall, these results highlight the strength of our method in challenging ill-exposed regimes.

For the X2HDR baseline, we adopt their direct HDR-generation formulation with Stable Diffusion~2.1 (SD~2.1) as the image-diffusion backbone. Following X2HDR, we train the model to predict PU-21-encoded HDR latents. We use SD~2.1 because it shares the same backbone family as Stable Video Diffusion, enabling a controlled comparison. Since X2HDR uses a Flux-based backbone~\citet{flux2024}, this SD~2.1 variant helps isolate the effect of the reconstruction formulation from differences in backbone capacity. We report the Flux-based X2HDR results in \autoref{app:x2hdr-flux}.
\paragraph{Discussion:}
Beyond the quantitative gains, generating an explicit exposure bracket gives our method an interpretable intermediate representation. Visualizing the generated bracket helps diagnose failure cases, such as insufficient dynamic range across exposures (\cref{fig:expbracket-viz}, row~3). The bracket also enables direct control over exposure spacing through inference-time optimization. Improving $\Delta EV$ consistency between adjacent frames (\cref{tab:inference-time-opt}, \cref{para:bracketcons}) makes the generated bracket compatible with fusion methods such as~\citet{debevec1997recovering}, which assume known exposure spacing.

Compared to LEDiff, which requires separate models for shadow and highlight recovery, we use a single video diffusion model to synthesize the full exposure sequence, covering both low- and high-exposure LDR images. In contrast to X2HDR, which directly predicts HDR latents in a PU-21-transformed space, our method generates an explicit LDR bracket, keeping the generative process in the image space where latent diffusion models are natively trained. 

\begin{table}
\centering
\small
\setlength{\tabcolsep}{2.5pt}
\begin{tabular}{l|cccccc}
\toprule
\textbf{Method} & \textbf{$Q^{\ast}$-PSNR $\uparrow$} & \textbf{$Q^{\ast}$-SSIM $\uparrow$} & \textbf{$Q^{\ast}$-MAE $\downarrow$} & \textbf{$Q^{\ast}$-LPIPS $\downarrow$} & \textbf{Q-JOD $\uparrow$} & \textbf{Time $\downarrow$} \\
\midrule
Ours & $\mathbf{22.36 \pm 3.48}$ & $0.664 \pm 0.156$ & $\mathbf{0.060 \pm 0.027}$ & $\mathbf{0.289 \pm 0.111}$ & $7.51 \pm 0.99$ & ${\sim}$15s \\
LEDiff & $18.97 \pm 3.07$ & $0.552 \pm 0.128$ & $0.098 \pm 0.043$ & $0.384 \pm 0.128$ & $6.55 \pm 0.98$ & ${\sim}$15s \\
BracketDiff & $21.41 \pm 3.87$ & $\mathbf{0.734 \pm 0.128}$ & $0.077 \pm 0.044$ & $0.353 \pm 0.161$ & $7.51 \pm 1.21$ & ${\sim}$6min \\
X2HDR-SD2.1 & $21.24 \pm 3.32$ & $0.585 \pm 0.150$ & $0.068 \pm 0.030$ & $0.335 \pm 0.120$ & $7.17 \pm 0.97$ & ${\sim}$15s \\
\midrule
HDRCNN & $21.50 \pm 6.27$ & $0.601 \pm 0.281$ & $0.098 \pm 0.073$ & $0.342 \pm 0.246$ & $\mathbf{7.66 \pm 1.68}$ & ${<}$1s \\
SingleHDR & $21.61 \pm 3.93$ & $0.698 \pm 0.176$ & $0.075 \pm 0.039$ & $0.302 \pm 0.180$ & $7.56 \pm 1.40$ & ${<}$1s \\
\bottomrule
\end{tabular}
\caption{\textbf{Quantitative HDR comparison.} Our method outperforms the generative baselines on most quantitative metrics (PSNR, MAE, LPIPS), while achieving competitive performance on the remaining metrics. We compare favorably with non-generative feedforward methods (HDRCNN, SingleHDR) while avoiding some of their failure modes in severely under- and over-exposed regions.}
\label{tab:hdr-comparison}

\end{table}
\paragraph{User Study:} We evaluate the qualitative preference rate of our method against the baselines through a user study. We display our output alongside a randomly chosen baseline from \cref{tab:hdr-comparison}, with all method labels hidden. We then show the ground-truth HDR image and ask the participant to select the image that best matches it. Each participant completes 96 randomized pairwise comparisons. Across 20 participants, our method achieves a $72\%$ preference rate in pairwise comparisons. A two-sided binomial test against chance ($p_0{=}0.5$) confirms a statistically significant preference for our method against every baseline ($p<10^{-4}$), as shown in \cref{tab:user-study}. We use a MacBook Pro with an HDR display and a peak brightness of $1600$ nits. See \autoref{app:H} for more details on the user-study setup. 
\begin{table}
\centering
\small
\setlength{\tabcolsep}{4pt}
\renewcommand{\arraystretch}{1.15}
\begin{tabular}{@{}l r r r r r r@{}}
\toprule
 & \textbf{LEDiff} & \textbf{HDRCNN} & \textbf{SingleHDR} & \textbf{BracketDiff} & \textbf{X2HDR-SD2.1} & \textbf{Overall} \\
\midrule
Win rate (Ours) $\uparrow$
& 86.2\% & 73.6\% & 71.5\% & 67.6\% & 60.8\% & \textbf{72.0\%} \\
Binomial \(p\) $\downarrow$
& $7.2{\times}10^{-51}$ & $1.1{\times}10^{-20}$ & $1.7{\times}10^{-17}$ & $4.6{\times}10^{-12}$ & $2.7{\times}10^{-5}$ & $\mathbf{2.8{\times}10^{-85}}$ \\
\bottomrule
\end{tabular}
\caption{\textbf{User-study results.} 
Participants compared our reconstructions against each baseline using the GT HDR
image as a reference. We report the percentage of trials preferring ours, with
\emph{Overall} aggregated over $1920$ pairwise comparisons. Two-sided exact
binomial tests against chance ($p_0{=}0.5$) show a significant preference rate for our method against the compared baselines.}
\label{tab:user-study}
\end{table}
\subsection{Ablation Studies}
\label{subsec:ablation}
\begin{figure}
    \centering
    \includegraphics[width=0.5\linewidth]{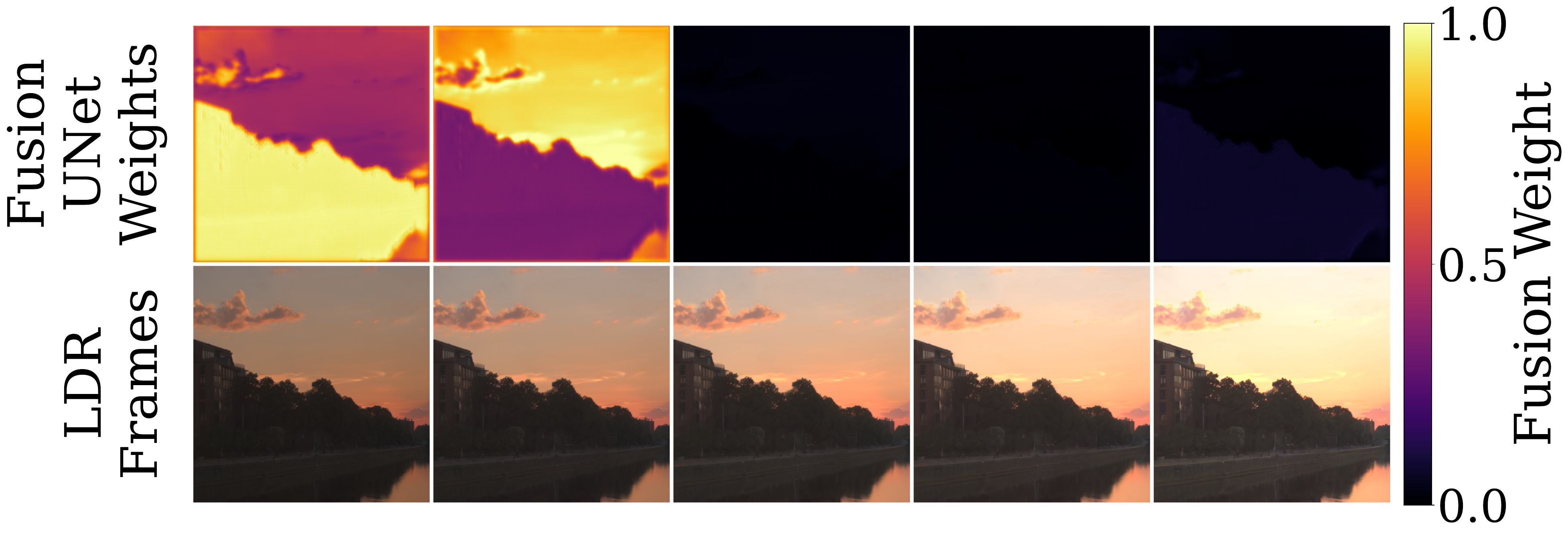}%
    \includegraphics[width=0.5\linewidth]{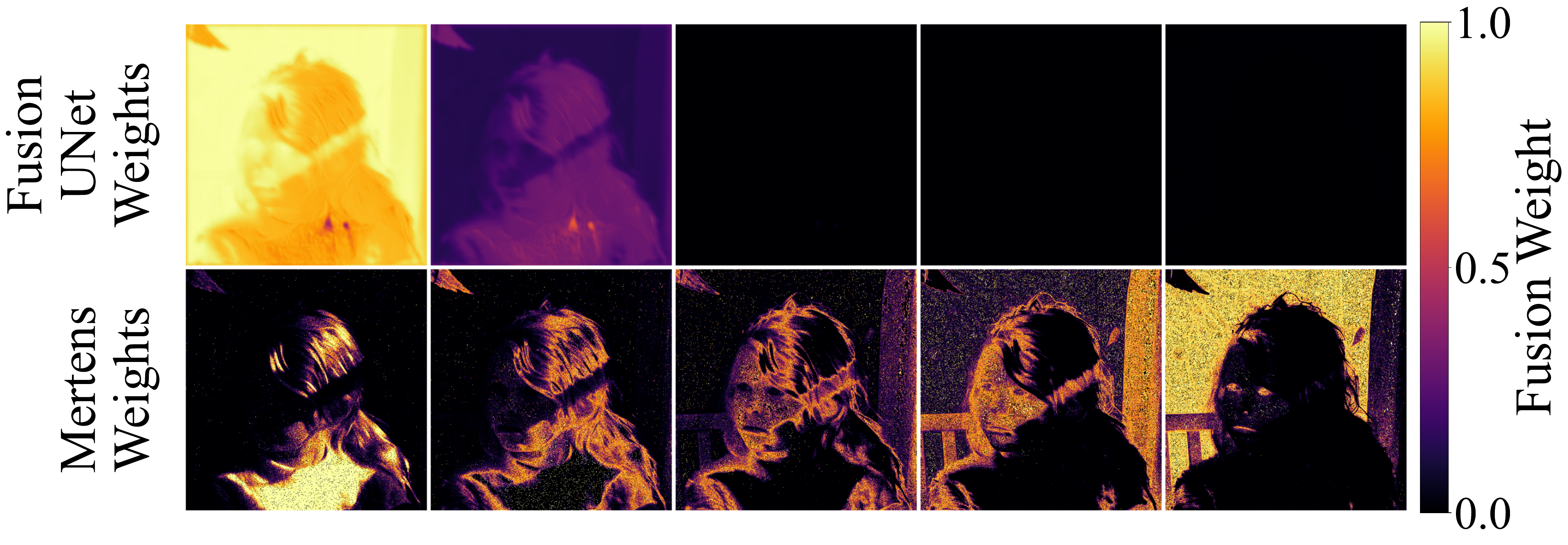}
    \caption{\textbf{Fusion method comparison.} (Left) UNet fusion weights on a test scene. (Right) Visual comparison of Mertens vs.\ UNet fusion on a test scene. Our Fusion UNet generates spatially smoother and less noisy fusion weights as compared to Mertens.}
    \label{fig:fusion-ablation-visual}
\end{figure}

\paragraph{Fusion strategy ablation:}
We ablate different strategies for fusing the generated LDR images by replacing our Fusion UNet with two widely used classical methods: Mertens fusion~\citet{mertens2007exposure} and Debevec merging~\citet{debevec1997recovering}. Measured over the test set, our trained Fusion UNet ($Q^{\ast}$-PSNR: 22.36) outperforms both Mertens fusion ($Q^{\ast}$-PSNR: 18.87) and Debevec merging ($Q^{\ast}$-PSNR: 15.23). All fusion methods receive the exposure brackets generated by the fine-tuned model as input. Debevec merging underperforms Mertens fusion because it relies on known relative exposure values, which are not predicted by our method. Our Fusion UNet is instead trained on LDR brackets synthesized from the GT exposure distribution used to fine-tune the video model, allowing it to learn a fusion strategy better suited to our predicted LDR brackets. It also produces spatially smoother weighting maps than Mertens fusion (see~\cref{fig:fusion-ablation-visual}). Please see \autoref{app:C} for more details.
\noindent
\begin{table}
\centering
\small
\begin{tabular}{ll|cccc}
\toprule
\textbf{Model} & \textbf{HDR Merge} & \textbf{$Q^{\ast}$-PSNR $\uparrow$} & \textbf{$Q^{\ast}$-SSIM $\uparrow$} & \textbf{$Q^{\ast}$-MAE $\downarrow$} & \textbf{$Q^{\ast}$-LPIPS $\downarrow$} \\
\midrule
Base Model & Debevec & 17.82 & 0.52 & 0.11 & 0.44 \\
\textbf{Ours} & \textbf{Debevec} & \textbf{21.36} & \textbf{0.62} & \textbf{0.06} & \textbf{0.34} \\
\bottomrule
\end{tabular}
\caption{\textbf{Comparison under inference-time optimization.} After enforcing exposure-consistent brackets before Debevec merging, our model outperforms the base SVD model.}
\label{tab:inference-time-opt}
\end{table} 
\paragraph{Bracket Consistency:}\label{para:bracketcons} While our current model is trained to generate monotonically increasing exposures, it does not explicitly enforce a fixed $\Delta EV$ between consecutive frames, where
$\Delta EV = \log_2(t_i/t_{i-1})$ denotes the log-ratio of their exposure times $(t_i, t_{i-1})$.
Known exposure spacing enables radiometrically consistent HDR recovery with methods such as Debevec merging. To achieve controlled exposure spacing, we further demonstrate that our method is compatible with inference-time optimization. 
Following BracketDiff, we optimize initial noise latents corresponding to the generated LDR bracket at inference time so that consecutive frames follow a monotonic $1$-EV spacing, i.e., $\Delta EV = 1$. \Cref{tab:inference-time-opt} shows that inference-time optimization is substantially more effective with our fine-tuned model than with the base Stable Video Diffusion (SVD) model.

We also achieve better bracket consistency between consecutive bracket images in comparison to the original SVD model after inference time optimization. We measure bracket consistency by comparing the relative brightness changes in the generated LDR bracket against the expected exposure spacing of a valid bracket. Since inference-time optimization enforces $\Delta EV = 1$ between adjacent LDR frames, corresponding well-exposed regions should exhibit an ideal brightness ratio of $2\times$. We therefore measure the mean absolute deviation of the observed adjacent-frame brightness ratios from the ideal value of $2$, where lower is better. Averaged over the test set, the fine-tuned model achieves a lower deviation ($0.233$) than the base SVD model ($0.326$), corresponding to an approximately $29\%$ improvement in bracket consistency. See \autoref{app:F} for details on bracket consistency evaluation. 

In \autoref{app:D}, we additionally report results for LoRA-based fine-tuning and using classifier-free guidance (CFG) during both fine-tuning and inference. In \autoref{app:E}, we show that reconstruction quality is primarily limited by the generated LDR bracket, rather than the Fusion UNet.
\subsection{Beyond Single-Shot HDR}
\label{subsec:focalstack}
\begin{figure}
    \centering
    \includegraphics[width=0.9\linewidth]{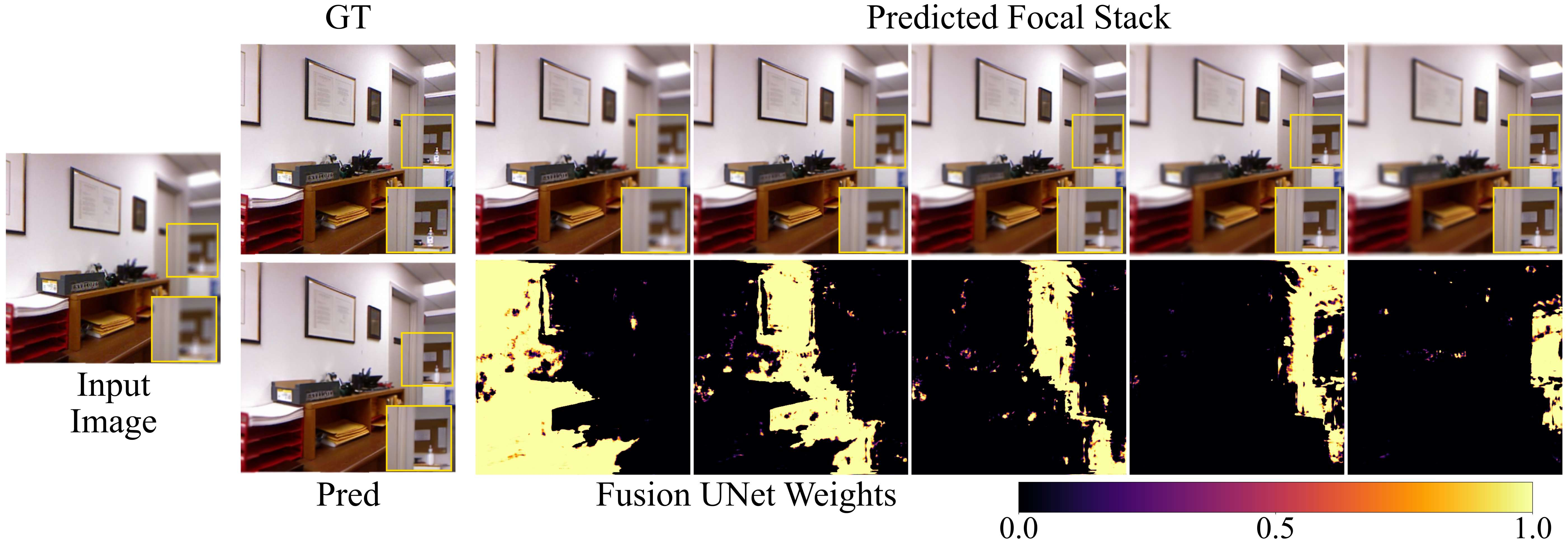}
    \caption{\textbf{Extending our method to all-in-focus (AIF) image recovery.}  We fine-tune the video model to take a defocused image as input and generate a focal stack of images with varying focus distances. We fuse the generated focal stack into an AIF image using a separate Fusion UNet.}
    \label{fig:defocus_results}
\end{figure}

Our framework naturally extends to other computational photography tasks. As a proof of concept, we demonstrate all-in-focus image recovery from a single defocus-blurred input in \cref{fig:defocus_results}. We fine-tune the video model to generate a focal stack with monotonically varying focus depth, analogous to the monotonic exposure progression in the HDR case. We then train a Fusion UNet on the ground-truth focal stacks to fuse them into an all-in-focus image. We simulate training data using RGB images and depth maps from the NYUv2 dataset~\citet{Silberman:ECCV12}, by rendering focal stacks with a simple depth-aware defocus blur model~\citet{gur2019single}. See \autoref{app:G} for more details. 
\section{Limitations and Future Work}
\vspace{-1mm}
\label{sec:limitations}
We recast single-shot HDR reconstruction as conditional video generation, where the generated frames form an exposure bracket that is fused into an HDR image using a light-weight network. This results in a simple and interpretable framework that is compatible with other computational photography tasks. However, our method is not without limitations. Scenes containing fine text or other high-frequency details (see \cref{fig:failure-cases}) can be degraded by the latent compression introduced by the Stable Video Diffusion VAE. In the trivial case, when the shadows and highlights are well captured in the input image, our method may slightly lag behind existing baselines. Since the input already preserves most of the scene content, such cases require minimal to no dynamic range expansion. We present a thorough analysis of these failure cases in \autoref{app:I}. These failure modes could be mitigated by training on more diverse datasets and stronger video generation backbones~\citet{wan2025wan}. Runtime can be substantially reduced by one-step distillation \citet{yin2024improved}, thereby bridging the gap between our approach and non-generative methods. Extending this framework to other reconstruction tasks, such as hyperspectral imaging, is a promising direction.

\paragraph{Acknowledgements:}
We thank Ishit Mehta and Ganesh Iyer for helpful feedback on the draft, Yash Belhe for suggesting useful experiments, and Xiang Dai for discussions on HDR imaging. We also gratefully acknowledge the Machine Intelligence, Computing, and Security (MICS) Center at UCSD for GPU and storage support. This work was supported in part by NASA under SBIR Award 318257-00001.
\begin{small}
\bibliographystyle{plainnat}
\bibliography{sample}
\end{small}
\newpage
\appendix
\section*{Appendix}
We organize the appendix as follows. In \autoref{app:A}, we describe the preprocessing used for the HDR training and evaluation datasets. \autoref{app:B} details our evaluation protocol, with emphasis on scale-invariant $Q^{\ast}$ metrics and the preprocessing applied before HDR-VDP3 evaluation. \autoref{app:C} provides additional quantitative results for the fusion ablation, comparing our Fusion UNet against classical LDR bracket merging methods. In \autoref{app:D}, we study the effect of classifier-free guidance and LoRA fine-tuning on the performance of our method. In \autoref{app:K}, we visualize the generated LDR brackets against the GT LDR brackets and also present a quantitative comparison. \autoref{app:E} analyzes the role of the Fusion UNet in the performance of our method, including the drop observed when applying it to LDR brackets produced after inference-time optimization of the video model. We define the bracket-consistency metric in \autoref{app:F}. \autoref{app:G} describes how our framework can be adapted to all-in-focus image estimation from a single defocus-blurred input. \autoref{app:H} reports additional details on the user-study setup. \autoref{app:I} presents a detailed failure-case analysis, showing when our method performs worse and how this relates to input edge density and high-frequency detail loss from the VAE. Finally, \autoref{app:x2hdr-flux} compares our method with X2HDR-Flux, which uses a substantially larger denoising backbone than our model.
 
\section{Training and Evaluation Dataset details}
\label{app:A}
We collate $10{,}965$ training crops ($5{,}820$ from Fairchild HDRPS~\citet{fairchild2007hdr},
$5{,}145$ from RawHDR~\citet{zou2023rawhdr}) using a recipe similar to
X2HDR~\citet{wu2026x2hdr}. Source images are first downsampled to a width of 2560\,px while preserving the aspect ratio. From each downsampled image, we extract 15 non-overlapping $512{\times}512$ crops, where the centers of the crops are selected via greedy farthest-first traversal of the regular crop grid: starting at the top-left position, each subsequently chosen origin maximizes the minimum squared pixel-distance to all previously-chosen origins, yielding spatially diverse, non-redundant samples that cover the full image. Raw and HDR are cropped while preserving the pixel-level alignment between them.

\paragraph{Fairchild HDRPS~\citet{fairchild2007hdr}:} We
use all 44 scenes (388 bracketed NEF exposures total).
The LDR exposure bracket for each scene is fused into a single ground-truth HDR using
homography alignment and a minimum-spanning-tree exposure
estimator~\citet{hanji2023robust} for a globally consistent radiometric
scale. Every NEF in the bracket is demosaiced to linear sRGB and
paired with the fused HDR image, giving one \texttt{(raw, gt\_hdr)} sample
per exposure ($388 \times 15 = 5\,820$ crops). This is a public domain dataset listed for non-commercial research.

\paragraph{RawHDR~\citet{zou2023rawhdr}:} We process 343 out of
345 packed-Bayer captures across the train and test splits
(two failed to load), each providing a short-exposure raw input and a
paired HDR raw. Both arrays are demosaiced and converted to linear
sRGB using the per-capture white-balance gains and $3{\times}3$
camera-to-sRGB matrix exported with the dataset, yielding $343 \times
15 = 5\,145$ pairs. This dataset is MIT licensed, which allows using it for research.

\paragraph{Test set:} Evaluation is performed on the
SI-HDR~\citet{hanji2022comparison} \texttt{raw2hdr} test set on 96 scenes, processed and provided by \citet{wu2026x2hdr} to ensure consistent evaluation between both methods. This dataset is released under a CC BY 4.0 license, and thus can be used for research purposes.

\paragraph{Fusion UNet training data:} Unlike the video diffusion model, which is conditioned on raw LDR inputs, the Fusion UNet only requires (clean) ground-truth HDR images at training time, since the input LDR bracket to the Fusion UNet is synthesized from the GT HDR image via \cref{eq:HDR2LDR}. This allows us to train the Fusion UNet using both the ground-truth HDR images from the dataset described above and a subset of the public HDR datasets used to train LEDiff~\citet{wang2025lediff}, restricted to sources that overlap with our collation. The combined corpus contains approximately $21{,}000$ HDR images.

\section{Evaluation Metrics Details}
\label{app:B}
Our method, similar to existing works~\citet{wang2025lediff, wu2026x2hdr, bemana2025bracket}, predicts the radiance up to a global scale. We use the $Q^{\ast}$ metrics proposed in~\citet{cao2024perceptual}, and also used in~\citet{wu2026x2hdr}, to ensure invariance to the global scales in the quantitative evaluation. Computing $Q^{\ast}$ metrics begins by optimizing the exposure values (using gradient descent) to match the GT and predicted HDR images based on a metric of choice, such as PSNR, SSIM, etc. Additionally, the evaluation averages spatially weighted metrics across a range of exposures, restricting each metric to image regions that are meaningful at the corresponding exposure. Applying the exposure mask is straightforward for pixel and patch-wise metrics such as SSIM, PSNR, and MAE. For LPIPS, we instead compute patch-wise perceptual distances first and apply the mask afterward, avoiding artifacts introduced by evaluating LPIPS on masked images. We omit DISTS~\citet{dists} as it is a global image-level metric, making a masked patch-wise evaluation nontrivial. Before computing the Q-JOD metric according to HDR-VDP3~\citet{mantiuk2023hdr}, we normalize the ground truth and baseline images to the same median luminance value (8.0), as Q-JOD uses the full global scale for evaluation. Prior work~\citet{wu2026x2hdr} also uses a similar normalization for displaying images in its user study. 

\section{Fusion Ablation Study Metrics}
\label{app:C}
We report the additional metrics in \cref{tab:fusion-ablation} to compare our Fusion UNet with the classical LDR bracket fusion methods outlined in \cref{subsec:ablation}. We apply the Mertens algorithm on linearized input brackets and treat the weighted reconstruction as a linear-domain estimate. This ensures that the output is evaluated in the same linear HDR domain as our Fusion UNet prediction and the ground-truth HDR image. Debevec merging requires the relative exposure differences between the input images. Since our video model predicts only the bracket images and not their corresponding exposure values, we assume a fixed synthetic exposure schedule by uniformly sampling EV offsets in the range $[-5.5, 5.5]$ across the generated bracket. This comparison, therefore, evaluates whether classical fusion methods can be used as drop-in replacements for our learned Fusion UNet under the same generated-bracket setting. The weaker performance of Debevec merging reflects its sensitivity to accurate exposure metadata, whereas our Fusion UNet does not require exposure estimates at inference time.
\begin{table}
\centering
\small
\begin{tabular}{l|cccc}
\toprule
\textbf{Fusion Method} & \textbf{Q*-PSNR $\uparrow$} & \textbf{Q*-SSIM $\uparrow$} & \textbf{Q*-MAE $\downarrow$} & \textbf{Q*-LPIPS $\downarrow$} \\
\midrule
Mertens~\citet{mertens2007exposure}      & 18.87 & 0.513 & 0.096 & 0.352 \\
Debevec~\citet{debevec1997recovering}      & 15.23 & 0.503 & 0.141 & 0.433 \\
\textbf{Ours (Fusion UNet)}     & \textbf{22.36} & \textbf{0.664} & \textbf{0.060} & \textbf{0.289} \\
\bottomrule
\end{tabular}
\caption{\textbf{Ablation study with different fusion methods.} Our learned Fusion UNet, trained on the same distribution of generated frames, substantially outperforms both the classical alternatives across all the metrics.}
\label{tab:fusion-ablation}
\end{table}

\section{Additional results from LoRA fine-tuning and no classifier-free guidance}
\label{app:D}
In~\cref{tab:ablation-cfg-lora} we report the metrics from a sweep of the CFG weight, showing that increasing the CFG scale slightly hurts performance. For this ablation, we used a model trained for $15{,}000$ iterations with conditional dropout enabled during training, following~\citet{chen2025repurposing}. We also observe slight performance drops on using LoRA fine-tuning (rank 128), where the LoRA layers were applied to the Q, K, V projections and the attention output blocks of the transformer layers of the Video Diffusion UNet. 
\begin{table}
\centering
\small
\begin{tabular}{l|ccccc}
\toprule
\textbf{Variant} & \textbf{Q$^\ast$-PSNR $\uparrow$} & \textbf{Q$^\ast$-SSIM $\uparrow$} & \textbf{Q$^\ast$-MAE $\downarrow$} & \textbf{Q$^\ast$-LPIPS $\downarrow$} & \textbf{Q-JOD $\uparrow$} \\
\midrule
\multicolumn{6}{l}{\emph{CFG sweep}} \\
CFG = 1.0 (no guidance)  & 22.33 & 0.654 & 0.060 & 0.293 & 7.14 \\
CFG = 1.5                & 22.26 & 0.655 & 0.061 & 0.294 & 7.14 \\
CFG = 2.5                & 22.13 & 0.656 & 0.062 & 0.295 & 7.13 \\
CFG = 4.0                & 21.88 & 0.655 & 0.064 & 0.300 & 7.12 \\
\midrule
\multicolumn{6}{l}{\emph{LoRA vs full fine-tune (no CFG)}} \\
LoRA                       & 22.00 & 0.656 & 0.063 & 0.296 & 7.10 \\
\textbf{Full FT} & \textbf{22.36} & \textbf{0.664} & \textbf{0.060} & \textbf{0.289} & \textbf{7.51} \\
\bottomrule
\end{tabular}
\caption{\textbf{Ablations on classifier-free guidance scale and LoRA vs full fine-tuning.}
Increasing CFG monotonically degrades most fidelity metrics (PSNR, MAE, LPIPS, JOD) for a model trained with condition dropout. LoRA fine-tuning (with no condition dropout) marginally underperforms full fine-tuning, suggesting that full-rank capacity benefits learning the HDR-bracket distribution.}
\label{tab:ablation-cfg-lora}
\end{table}

\section{Generated exposure brackets from the video diffusion model}
\label{app:K}

\begin{table*}
  \centering
  \small
  \setlength{\tabcolsep}{5pt}
  \begin{tabular}{c c c c c c c c}
    \toprule
    & \multicolumn{3}{c}{Original prediction}
    & \multicolumn{4}{c}{After exposure compensation} \\
    \cmidrule(lr){2-4}
    \cmidrule(lr){5-8}
    Bracket idx
    & PSNR$\,\uparrow$ & SSIM$\,\uparrow$ & LPIPS$\,\downarrow$
    & PSNR$\,\uparrow$ & SSIM$\,\uparrow$ & LPIPS$\,\downarrow$
    & $\Delta\text{EV}$ \\
    \midrule
    0 (darkest)   & 20.76 & 0.545 & 0.270 & 32.53 & 0.838 & 0.116 & $-2.45$ \\
    1             & 18.31 & 0.591 & 0.263 & 27.38 & 0.810 & 0.151 & $-1.72$ \\
    2             & 17.25 & 0.652 & 0.262 & 23.16 & 0.768 & 0.204 & $-0.97$ \\
    3             & 17.60 & 0.681 & 0.260 & 20.74 & 0.713 & 0.251 & $-0.41$ \\
    4 (brightest) & 21.45 & 0.693 & 0.266 & 21.67 & 0.694 & 0.267 & $-0.03$ \\
    \midrule
    All           & \textbf{19.07} & \textbf{0.632} & \textbf{0.264}
                  & \textbf{25.10} & \textbf{0.765} & \textbf{0.198}
                  & $-1.12$ \\
    \bottomrule
  \end{tabular}
  \caption{Per-bracket-index reconstruction metrics between predicted and GT LDR brackets on the SI-HDR test set, before and after per-frame global-scale exposure compensation using \cref{eq:scale_alignment}. $\Delta\text{EV}=\log_2 s^\star$ denotes the optimal exposure offset applied to the prediction. Exposure compensation improves PSNR by $+6.0$\,dB on average and by $+11.8$\,dB at the darkest index, suggesting that much of the error comes from global EV miscalibration rather than structural mismatch.}
  \label{tab:ldr_bracket_metrics}
\end{table*}

\begin{figure}
    \centering
    \includegraphics[width=\linewidth]{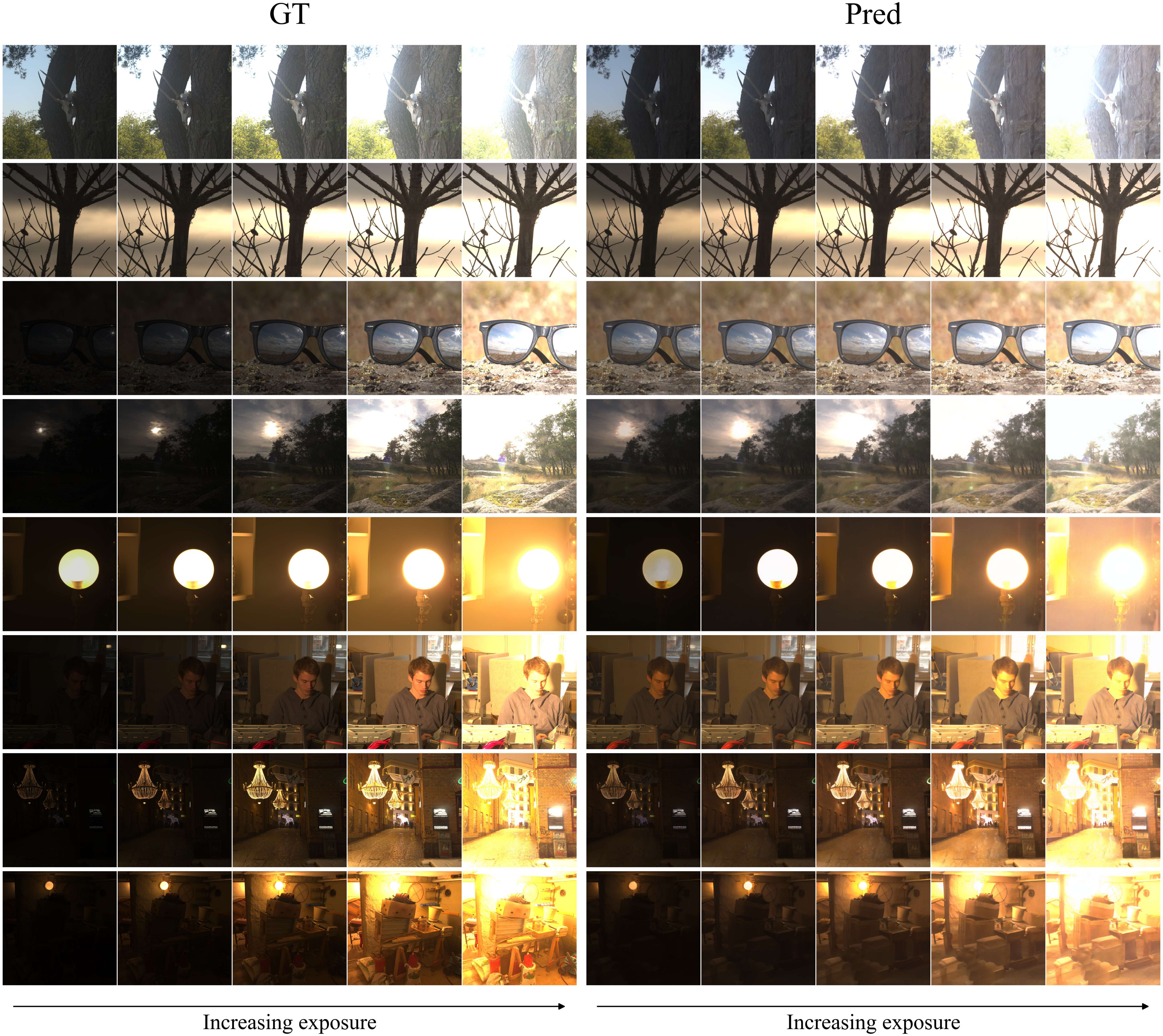}
    \caption{\textbf{Exposure brackets generated by our fine-tuned model.}
    Each row corresponds to one scene from the SI-HDR test set. We show the ground-truth LDR exposure bracket under the header \textit{GT} and our predictions under \textit{Pred}. Exposure increases from left to right under each header. Our predictions match the relative exposure progression of the GT brackets with high fidelity in most cases. In rows 3, 4, and 6, we observe an exposure mismatch at the low-EV (darkest) end of the bracket between our method and GT. However, this behavior is not consistent across the visualization, indicating output variance of the model rather than a systematic bias.}
    \label{fig:expbracket-viz}
\end{figure}

We visualize the LDR brackets synthesized by our method in \cref{fig:expbracket-viz}, alongside GT LDR brackets generated from the HDR images using \cref{eq:HDR2LDR} and \cref{eq:expselection}. Recall that the video model is fine-tuned on the training data to reproduce the exposure brackets generated by \cref{eq:HDR2LDR} and \cref{eq:expselection}. The synthesized brackets largely match the GT exposure progression, although in some cases the generated frames deviate in exposure, especially near the low-exposure end of the bracket.

We show in \cref{tab:ldr_bracket_metrics} that these errors are largely photometric rather than structural. Specifically, much of the discrepancy can be explained by a global per-frame exposure offset in linear light. Let $I_p, I_g \in [0,1]^{H \times W \times 3}$ denote a predicted LDR frame and its GT counterpart, both gamma-encoded with $\gamma=2.2$. We first linearize them as
\[
    \tilde{I}_p = I_p^\gamma, 
    \qquad 
    \tilde{I}_g = I_g^\gamma .
\]
We then solve for the optimal global linear-light scale
\begin{equation}
    s^\star
    =
    \argmin_{s > 0}
    \left\lVert s \tilde{I}_p - \tilde{I}_g \right\rVert_2^2
    =
    \frac{\langle \tilde{I}_p, \tilde{I}_g \rangle}
         {\langle \tilde{I}_p, \tilde{I}_p \rangle},
    \label{eq:scale_alignment}
\end{equation}
where inner products are computed over all pixels and color channels. The exposure-aligned prediction is
\[
    \hat{I}_p =
    \mathrm{clamp}\!\left(s^\star \tilde{I}_p, 0, 1\right)^{1/\gamma}.
\]
We then recompute PSNR, SSIM, and LPIPS between $\hat{I}_p$ and $I_g$. After this simple exposure alignment, the metrics improve substantially, especially for the darkest bracket frames. This indicates that the generated brackets preserve much of the underlying structure, and that the dominant mismatch is often a global EV miscalibration. Importantly, such global scale offsets do not directly affect downstream HDR reconstruction, since HDR predictions are evaluated up to an unknown global scale.

\section{Performance ceiling from Fusion UNet}
\label{app:E}
\begin{table}
\centering
\small
\begin{tabular}{p{0.38\linewidth}|cccc}
\toprule
\textbf{Method} & \textbf{Q*-PSNR $\uparrow$} & \textbf{Q*-SSIM $\uparrow$} & \textbf{Q*-MAE $\downarrow$} & \textbf{Q*-LPIPS $\downarrow$} \\
\midrule
Ours (Pred LDR brackets + Fusion UNet)       & 22.36 & 0.664 & 0.060 & 0.289 \\
Ours (Pred LDR brackets + exp comp + Fusion UNet) & 22.28 & 0.657 & 0.061 & 0.294 \\
\textbf{Fusion UNet + GT LDRs (Oracle)} & \textbf{44.84} & \textbf{0.989} & \textbf{0.007} & \textbf{0.013} \\
\bottomrule
\end{tabular}
\caption{\textbf{Fusion UNet oracle analysis.}
Ground-truth LDR brackets yield much higher scores (row 3) than our method (row 1), showing that the Fusion UNet is not the primary bottleneck. Exposure-compensating predicted brackets (row 2) has a negligible effect, indicating that global EV drift is not the limiting factor; performance is mainly limited by artifacts and inconsistencies in the generated LDR brackets.}
\label{tab:fusion-unet-oracle}
\end{table}

The performance of our pipeline depends on both the video diffusion model and the Fusion UNet. To isolate the effect of the Fusion UNet, we feed it ground-truth LDR brackets at test time. As shown in \cref{tab:fusion-unet-oracle}, this oracle setting substantially improves performance ($Q^{\ast}$-PSNR 44.84 vs. 22.36), indicating that the Fusion UNet is not the primary bottleneck. We also test whether $EV$ drift towards lower exposures in our predicted brackets affects downstream reconstruction by applying the scale alignment from \cref{eq:scale_alignment} before fusion. The negligible change in row 2 of \cref{tab:fusion-unet-oracle} shows that exposure drift is not the limiting factor. Instead, performance is mainly limited by artifacts and inconsistencies in the generated LDR brackets, suggesting that improvements to the video model should directly improve HDR reconstruction.
However, the Fusion UNet underperforms on brackets produced after inference-time optimization (\cref{subsec:ablation}), where the video model is constrained to generate fixed $\Delta EV{=}1$ exposure spacing. In this setting, Debevec merging performs better. We attribute this drop to distribution shift: the Fusion UNet is trained on GT brackets from \cref{eq:HDR2LDR} and \cref{eq:expselection}, whose $\Delta EV$ spacing is input-dependent, but is tested on brackets modified to have constant spacing. This performance drop is hence expected as the Fusion UNet is intentionally lightweight and not designed to handle arbitrary bracket distributions. Its role is to show that generated LDR brackets can support HDR reconstruction through simple learned per-pixel fusion. Quantitative results are provided in \cref{tab:inference-unet-drop}.
\begin{table}
\centering
\small
\begin{tabular}{l|cccc}
\toprule
\textbf{HDR Merge} & \textbf{$Q^{\ast}$-PSNR $\uparrow$} & \textbf{$Q^{\ast}$-SSIM $\uparrow$} & \textbf{$Q^{\ast}$-MAE $\downarrow$} & \textbf{$Q^{\ast}$-LPIPS $\downarrow$} \\
\midrule
\textbf{Debevec} & \textbf{21.36} & \textbf{0.623} & \textbf{0.064} & \textbf{0.337} \\
Fusion UNet          & 16.20          & 0.570          & 0.127          & 0.431          \\
\bottomrule
\end{tabular}
\caption{\textbf{Effect of distribution mismatch on Fusion UNet.} After inference-time optimization, Debevec merging outperforms the Fusion UNet. The optimized LDR brackets induce a distribution shift relative to the brackets used during Fusion UNet training and negatively affect its performance.}
\label{tab:inference-unet-drop}
\end{table}
\section{Bracket Consistency Experiment Details}
\label{app:F}
To evaluate whether the generated LDR frames form a physically plausible exposure bracket, we measure bracket consistency on 96 scenes from the SI-HDR test set. A bracket-consistent sequence should exhibit monotonically increasing linear brightness with adjacent frames separated by one exposure stop ($\Delta EV =1)$, i.e., a $2\times$ increase in linear intensity. 
For each generated sequence, we first linearize the frames using the inverse sRGB gamma approximation, $x^{2.2}$, and compute the luminance image. To avoid confounding the measurement with saturation or clipping, we compute a shared valid mask for each generated bracket using the linearized luminance values. Specifically, we retain only pixels $p$ that remain unsaturated in all five generated frames:
\[
    0.04 < Y_i(p) < 0.96, \qquad \forall i \in \{1,\ldots,5\},
\]
where $Y_i(p)$ denotes the linearized luminance at pixel $p$ in frame $i$. Bracket consistency is then measured only over this shared valid region.

Here, we outline the quantitative computation of bracket consistency as reported in \cref{subsec:ablation}. We evaluate bracket consistency locally over non-overlapping $64\times64$ patches. A patch is retained only if at least $50\%$ of its pixels are valid under the shared mask. For scene $s$, valid patch $p$, and exposure index $k$, let $\ell_{s,p,k}$ denote the mean linear luminance of the $k$-th generated frame over the valid pixels in that patch. We compute the brightness ratio between adjacent exposure steps as
\[
r_{s,p,k} = \frac{\ell_{s,p,k+1}}{\ell_{s,p,k} + \epsilon},
\]
where $k$ indexes the four adjacent transitions in the five-frame bracket. We define the ratio deviation as
\[
\delta =
\frac{1}{|\mathcal{R}|}
\sum_{(s,p,k)\in \mathcal{R}}
\left| r_{s,p,k} - 2.0 \right|,
\]
where $\mathcal{R}$ is the set of all valid scene--patch--transition tuples. Thus, lower values indicate more consistent exposure spacing, with $\delta=0$ corresponding to perfect $2\times$ spacing at every measured transition. The bracket-consistency numbers reported in \cref{subsec:ablation} are computed using the $\delta$ metric defined here. 
\section{Extending our approach to all-in-focus image estimation}
\label{app:G}
We provide details for adapting our approach to all-in-focus (AIF) image estimation, as discussed in \cref{subsec:focalstack}.

\paragraph{Focal stacks and all-in-focus estimation:}
Many computer vision pipelines assume a pinhole camera model, where the captured image is sharp everywhere. Such an image is commonly referred to as an all-in-focus (AIF) image. In contrast, real cameras have finite apertures, which introduce depth-dependent defocus blur. This blur can be modeled using geometric optics~\citet{potmesil1981lens}. A common approximation is to synthesize a defocused image $I_b$ from an AIF image $I$ using a spatially varying blur kernel:
\begin{equation}
    I_b(x,y) = \int\int I(u,v) \, h(x,y,u,v \mid D(u,v), F) \, du\,dv,
\end{equation}
where $(x,y)$ denotes the output pixel location, $(u,v)$ denotes coordinates over the image domain, $D(u,v)$ is the scene depth, and $F$ is the focus distance. For a fixed camera aperture, the blur kernel depends on the mismatch between the scene depth and the focus distance. The blur magnitude increases with $|D(u,v)-F|$, according to the circle-of-confusion equation (Eq.1 of~\citet{gur2019single}). Please refer to ~\citet{gur2019single} for a detailed overview of the defocus blur forward model.

\paragraph{Recovering AIF images from a focal stack:}
Different parts of a scene are in focus (blur-free) at different focus distances. Let $\{I_b^1,\dots,I_b^N\}$ denote a focal stack captured with monotonically increasing focus distances $\{F^1,\dots,F^N\}$. Conventional AIF recovery methods~\citet{li2013image, lee2016robust} estimate per-pixel weight maps $\{W^1,\dots,W^N\}$ over the stack and recover the AIF image as
\begin{equation}
    I_{\mathrm{AIF}} = \sum_i W^i \odot I_b^i.
\end{equation}
However, these methods require capturing a multi-image focal stack as input. Under our proposed paradigm, we instead fine-tune the video model to generate a focal stack with monotonically increasing focus distances, conditioned on a single input image. A light-weight fusion network then predicts per-pixel fusion weights and combines the generated stack into an AIF estimate.

\paragraph{Experiment details:}
We evaluate this adaptation on the NYUv2 dataset~\citet{Silberman:ECCV12}. We use 1250 RGB-D images for training. For each RGB image and its corresponding depth map, we simulate a focal stack of five defocused images using the differentiable defocus blur approximation from prior work~\citet{gur2019single}. To choose the focus distances for each scene, we run K-means clustering on the ground-truth depth map with $K=5$, selecting focus distances that cover the dominant scene depths.

During fine-tuning, a conditioning input image is randomly sampled from the simulated focal stack. The video diffusion model is fine-tuned conditioned on the input to generate the full focal stack, ordered monotonically by focus distance. In a separate training stage, we then train the Fusion UNet on generated GT focal stacks to predict per-pixel fusion weights, using an $\ell_1$ reconstruction loss against the original sharp RGB images, which we treat as the ground-truth AIF targets. Thus, the overall pipeline is identical to our HDR reconstruction method in \cref{sec:method}, but applied to a different image formation model.

Averaged over 50 randomly selected scenes from the NYUv2 test set, our method achieves a PSNR of $31.26$, SSIM of $0.90$, and LPIPS of $0.096$ against the original sharp RGB images. These results suggest that our formulation is not specific to exposure bracketing and can extend to other inverse problems where the desired image can be recovered by synthesizing and fusing a physically meaningful image stack.

\section{Details on User Study Setup}
\label{app:H}
\paragraph{Apparatus:} All sessions were run on a 14'' MacBook Pro with the built-in Liquid Retina XDR display (peak 1600 nits, EDR-capable), under standard office lighting that was kept consistent across participants. The viewer was a custom Swift/Metal application that operates in a linear color space and writes into an extended-dynamic-range Float16 layer, so values above SDR reference white are preserved through to the HDR headroom of the display, where HDR headroom is defined as the ratio of the peak luminance to the SDR reference white value.

\paragraph{Stimuli and pre-display pipeline:} Each trial draws on a single scene from the test split, where the same scene was reconstructed by every method (ours and the five baselines: LEDiff, HDRCNN, SingleHDR, BracketDiff, and X2HDR-SD2.1) plus a ground-truth HDR image. To prevent the choice of normalization from biasing preferences, every image is sent through a fixed pipeline before display: (i) we use the linear HDR images directly; (ii) we normalize each image by dividing by its own $20.05^{\mathrm{th}}$-percentile luminance and multiplying by an exposure constant of $0.0037$, anchoring the dark end of the histogram to a common level; (iii) the result is scaled by a fixed EDR gain of $12\times$ to push the working range into the HDR headroom of the display. The ground-truth image is rendered through the same pipeline, with no preferential treatment.

\begin{figure}
    \centering
    \includegraphics[width=0.6\linewidth]{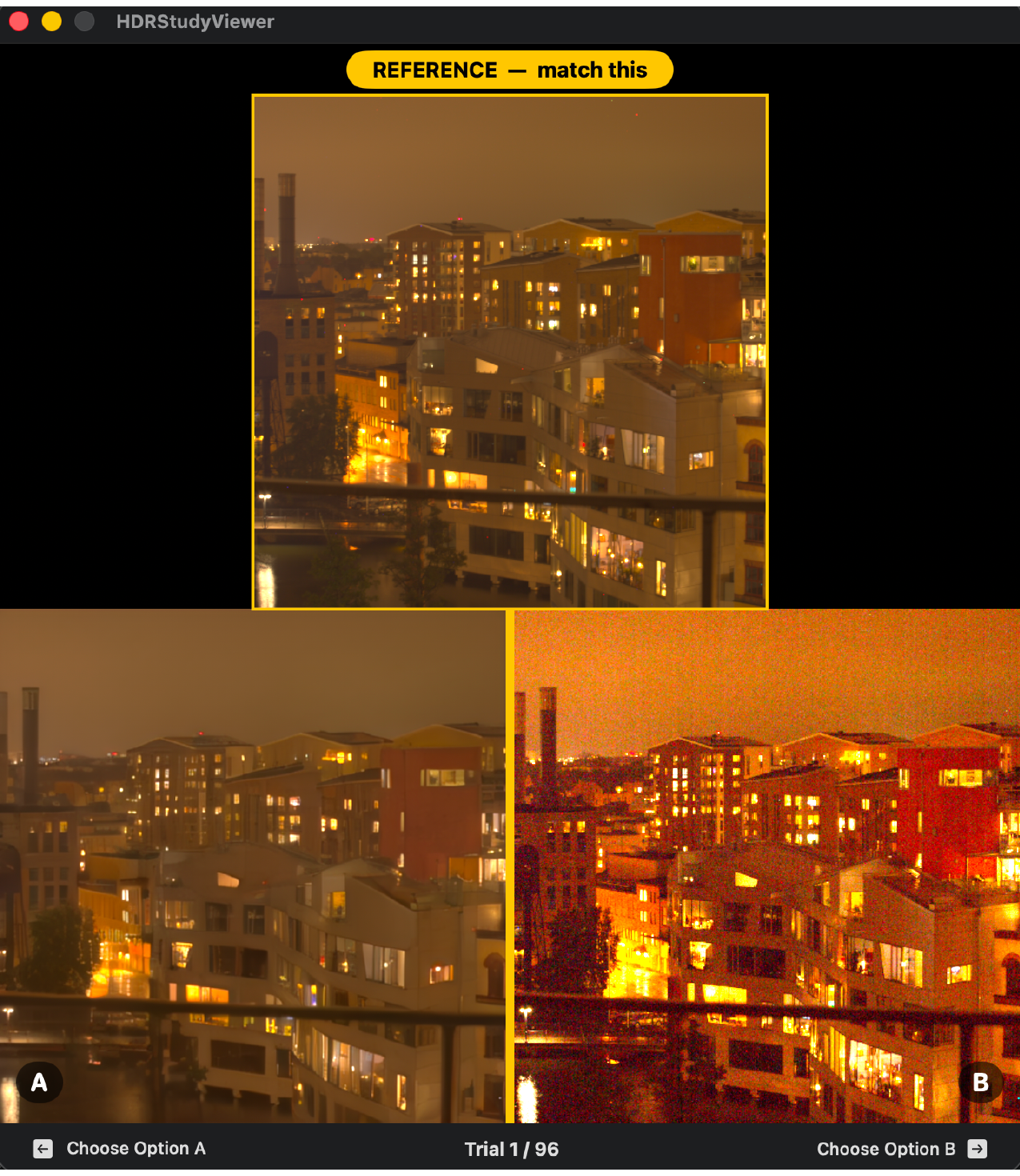}
    \caption{\textbf{User-study trial interface.} The ground-truth HDR image is displayed on the top half (centered) and the two candidates (ours vs.\ a randomly chosen baseline) are displayed side-by-side on the bottom half. The participant uses the left/right arrow keys to pick the candidate that best matches the ground truth.}
    \label{fig:user-study-ui}
\end{figure}

\paragraph{Task and protocol:} We use a three-image layout with the ground truth on the top half (centered) and the two candidates (ours vs.\ a randomly chosen baseline) side-by-side on the bottom half; left/right placement is randomized per trial. The participant chooses the bottom image that best matches the ground truth using the left/right arrow keys; \cref{fig:user-study-ui} shows the on-screen arrangement. There is no time limit. Each participant completes 96 trials drawn such that each baseline appears with approximately equal frequency, with scene order randomized per participant. The user study was uncompensated.

\paragraph{Participants:} We collected data from $N{=}20$ participants (14 male, 4 female, 2 non-binary), ages $18$--$58$ (median $22.5$, mean $26.2$); $10$ wore glasses or contacts. Sessions lasted approximately $10$--$30$ minutes (median $\approx 22$ min). Demographics were recorded with consent.

\paragraph{Aggregated results:} \cref{tab:user-study} reports the per-baseline ours-preferred rate; the overall rate is $72.0\%$ ($1382/1920$ judgements). We observe a minor difference in preference rates when our method was displayed in the left pane ($70.7\%$) vs.\ the right pane ($73.3\%$), indicating that the study was not affected by any image-placement bias. Over all pairwise comparisons, our method was displayed in the left and right panes with almost uniform probability ($49.0\%$ vs.\ $51.0\%$). Per-participant rates ranged from $58.3\%$ to $81.2\%$, with every participant preferring our reconstruction in aggregate.

\paragraph{Statistical significance:} Following \citet{wang2025lediff}, we test each baseline with a two-sided binomial test against chance ($H_0\!:p{=}0.5$, where $p$ is the probability that a random participant prefers our reconstruction over the baseline). All five comparisons are significant well below the $\alpha{=}0.01$ level (\cref{tab:user-study}); the smallest margin is X2HDR-SD2.1 at $p=2.7\!\times\!10^{-5}$, and the pooled overall test gives $p=2.8\!\times\!10^{-85}$ on $1382/1920$ aggregate comparisons.

\paragraph{Per-participant analysis:} The trial-level binomial test in \cref{tab:user-study} treats all $1920$ judgements as independent, but the observations are clustered within $20$ participants ($96$ trials each), which can inflate apparent statistical confidence. We therefore report the raw per-participant preference rate against each baseline in \cref{tab:user-study-perparticipant}. Every one of the $20$ participants prefers our reconstruction in aggregate (range $58\%$–$81\%$), and a majority of participants also prefers our reconstruction against each baseline ($20/20$ for LEDiff and HDRCNN, $19/20$ for BracketDiff, $18/20$ for SingleHDR, and $15/20$ for X2HDR-SD2.1). The unanimity in the \emph{Overall} column yields a two-sided sign test of $p = 2(1/2)^{20} \approx 1.9\!\times\!10^{-6}$, which requires no distributional assumptions on the per-participant rates.

\begin{table}
\centering
\scriptsize
\setlength{\tabcolsep}{4pt}
\renewcommand{\arraystretch}{1.10}
\begin{tabular}{@{}l c c c c c c@{}}
\toprule
\textbf{Participant} & \textbf{LEDiff} & \textbf{HDRCNN} & \textbf{SingleHDR}
& \textbf{BracketDiff} & \textbf{X2HDR-SD2.1} & \textbf{Overall} \\
\midrule
P01 & 100\% & 77\% & 81\% & 86\% & 65\% & 81\% \\
P02 &  95\% & 72\% & 94\% & 81\% & 60\% & 80\% \\
P03 &  95\% & 75\% & 80\% & 76\% & 71\% & 80\% \\
P04 &  79\% & 92\% & 67\% & 89\% & 61\% & 78\% \\
P05 &  86\% & 74\% & 87\% & 65\% & 67\% & 77\% \\
P06 &  86\% & 85\% & 60\% & 68\% & 80\% & 77\% \\
P07 & 100\% & 73\% & 71\% & 54\% & 67\% & 74\% \\
P08 &  82\% & 76\% & 86\% & 74\% & 60\% & 74\% \\
P09 &  78\% & 94\% & 78\% & 59\% & 57\% & 73\% \\
P10 &  89\% & 59\% & 69\% & 74\% & 75\% & 73\% \\
P11 &  86\% & 71\% & 79\% & 82\% & 40\% & 73\% \\
P12 &  89\% & 63\% & 82\% & 57\% & 67\% & 71\% \\
P13 &  78\% & 74\% & 79\% & 62\% & 63\% & 71\% \\
P14 &  83\% & 61\% & 77\% & 84\% & 50\% & 71\% \\
P15 &  87\% & 54\% & 62\% & 65\% & 63\% & 68\% \\
P16 & 100\% & 65\% & 46\% & 79\% & 62\% & 67\% \\
P17 & 100\% & 67\% & 80\% & 54\% & 46\% & 66\% \\
P18 &  59\% & 85\% & 65\% & 69\% & 36\% & 66\% \\
P19 &  83\% & 70\% & 60\% & 37\% & 68\% & 62\% \\
P20 &  74\% & 75\% & 45\% & 52\% & 43\% & 58\% \\
\midrule
\textbf{\# >50\%} & \textbf{20/20} & \textbf{20/20} & \textbf{18/20}
& \textbf{19/20} & \textbf{15/20} & \textbf{20/20} \\
\bottomrule
\end{tabular}
\caption{\textbf{Per-participant aggregation of the user study.} Each
cell is the fraction of trials (out of approximately $20$ per
cell, varying with random scene assignment) in which that participant preferred our reconstruction over the
indicated baseline; the \emph{Overall} column is the same fraction
pooled across all five baselines ($96$ trials per participant). The
bottom row counts how many of the $20$ participants exceeded $50\%$ in
each column.}
\label{tab:user-study-perparticipant}
\end{table}
 
\begin{figure}
    \centering
    \includegraphics[width=\linewidth]{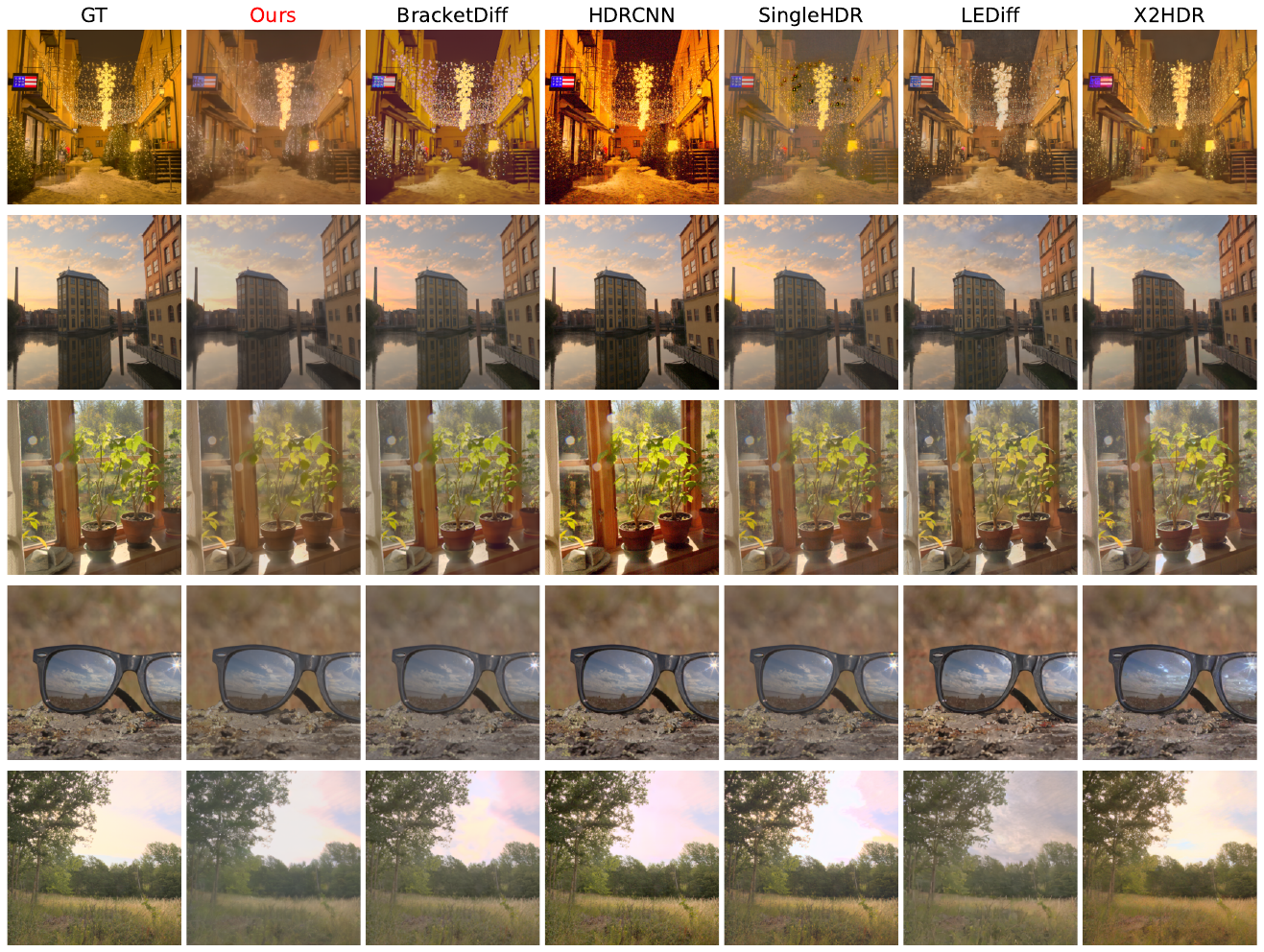}
    \caption{\textbf{User study failure cases.} Reconstructions for the five test-set scenes on which participants most often preferred a competing baseline over ours.}
    \label{fig:userstudy-worst}
\end{figure}

\paragraph{Qualitative worst cases (user study):} \cref{fig:userstudy-worst} shows the five test-set scenes on which our reconstructions were preferred least often by participants (preference rates ranging from $25\%$ to $40\%$ out of $20$ judgements per scene). These tend to be scenes with already-balanced exposure and salient mid-frequency texture (e.g.\ the potted plants in row 3), where dynamic-range expansion contributes little and any small detail loss from the latent VAE is more visible than on heavily ill-exposed inputs. 

\section{Failure case analysis}
\label{app:I}
\label{supp:failure-cases}
We expand on the limitations noted in \cref{sec:limitations} with a more thorough analysis. As noted in the main paper, our method can lag behind baseline methods on already well-exposed inputs, where little dynamic range
expansion is required. This limitation is less severe in practice, since such inputs already retain most of the visible scene content and benefit less from HDR reconstruction. To better understand this behavior, we analyze whether the
relative performance of our method is related to the amount of high-frequency details visible in the input. Intuitively, when fewer regions are saturated or under-exposed, more edges remain measurable in the input. Based on this insight, we use the \emph{edge density}, defined below, as a simple proxy for visible high-frequency scene detail. We find that this quantity strongly predicts the relative performance of our method against the baselines.

\paragraph{Edge density:}
For an input LDR image $I$ with luminance
$L = 0.2126\,R + 0.7152\,G + 0.0722\,B$, we compute the
$3{\times}3$ Sobel gradient magnitude
$G_{ij} = \sqrt{(L * S_x)_{ij}^2 + (L * S_y)_{ij}^2}$,
where $S_x$ and $S_y$ are the standard Sobel kernels~\citet{sobel19683x3}.
We define the edge density ($\kappa(\cdot)$) as the fraction of pixels with gradient magnitude above a threshold $\tau=0.13$:
\begin{equation}
\kappa(I) = \frac{1}{HW}\left|\{(i,j): G_{ij} > \tau\}\right|.
\end{equation}
We use $\kappa(I)$ as a simple proxy for the amount of visible high-frequency structure in the input LDR image.

\begin{figure}
    \centering
    \includegraphics[width=0.6\linewidth]{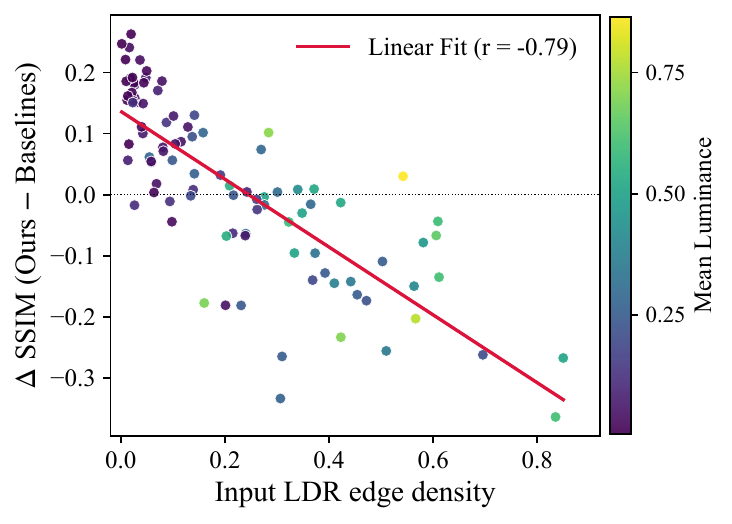}
    \caption{%
        \textbf{Per-scene $\Delta$\,SSIM vs.\ input edge density ($n=96$).}
        Each point is one test scene.
        $\Delta\,\text{SSIM} = \text{SSIM}_{\text{Ours}} - \overline{\text{SSIM}}_{\text{baselines}}$
        (positive $\Rightarrow$ Ours better).
        The Pearson correlation coefficient between edge density and $\Delta\,\text{SSIM}$
        is $r=-0.79$: We outperform the baselines on inputs with low edge-density. After controlling for the mean
        LDR luminance (normalized to $[0,1]$), edge density remains strongly correlated with $\Delta$\,SSIM (partial $r=-0.67$), while luminance contributes no
        residual signal once edge density is held fixed (partial $r=+0.04$). This suggests that the observed trend is driven primarily by the amount of visible structure in the input, rather than by its mean brightness.
    }
    \label{fig:edge_vs_delta_ssim}
\end{figure}

In \cref{fig:edge_vs_delta_ssim}, we plot $\Delta$\,SSIM against the input
edge density $\kappa(I)$ across all $96$ test scenes. For each scene,
$\Delta$\,SSIM is computed as the difference between the $Q^{\ast}$-SSIM of
our method and the average $Q^{\ast}$-SSIM of the baseline methods. We observe
from \cref{fig:edge_vs_delta_ssim} that edge density is a strong predictor of the relative performance gap between our method and the baselines. In particular, our method performs best relative to prior methods on low-edge-density inputs, which commonly correspond to
challenging HDR scenes with saturated or under-exposed regions.

The strong correlation in \cref{fig:edge_vs_delta_ssim} motivates a closer look
at what low edge density represents. On the SI-HDR test set, low input edge
density can occur either because the scene is intrinsically smooth, with little
high-frequency spatial content, or because the input LDR image is severely
under- or over-exposed, causing its limited dynamic range to suppress gradients
that would otherwise be visible. Both cases fall in the low-edge-density regime
where our method performs strongest relative to the baselines.

\begin{figure}
    \centering
    \includegraphics[width=\linewidth]{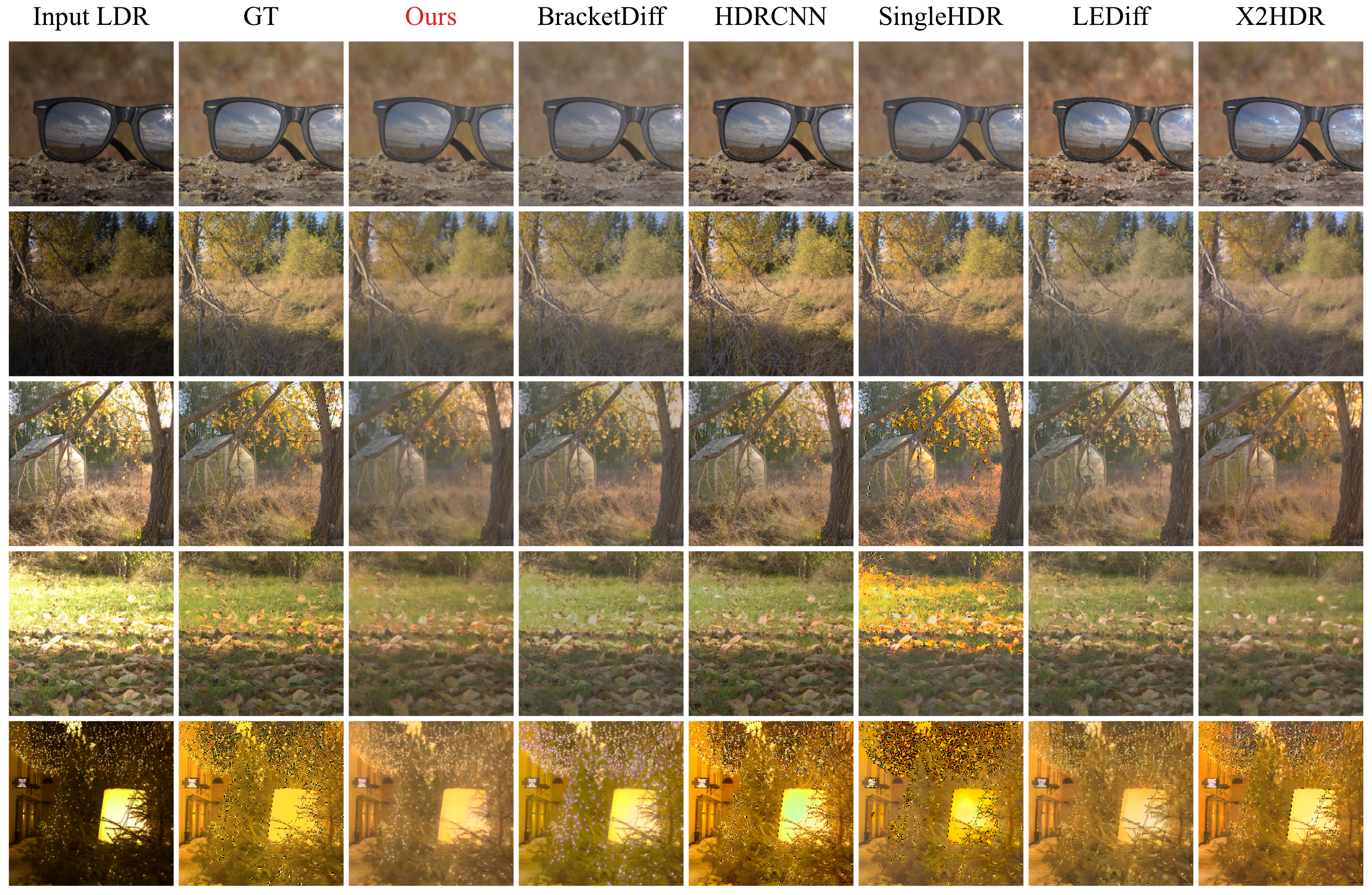}
\caption{\textbf{Failure-case qualitative comparison.}
We show five well-exposed scenes containing high-frequency textures. HDR images are visualized with Reinhard tone-mapping. In these cases, our method loses some fine texture detail compared to existing methods, consistent with the VAE bottleneck analysis in \cref{supp:vae-roundtrip}. Nevertheless, the reconstructions remain visually comparable overall.}
    \label{fig:worst-results}
\end{figure}

\begin{figure}
    \centering
    \includegraphics[width=\linewidth]{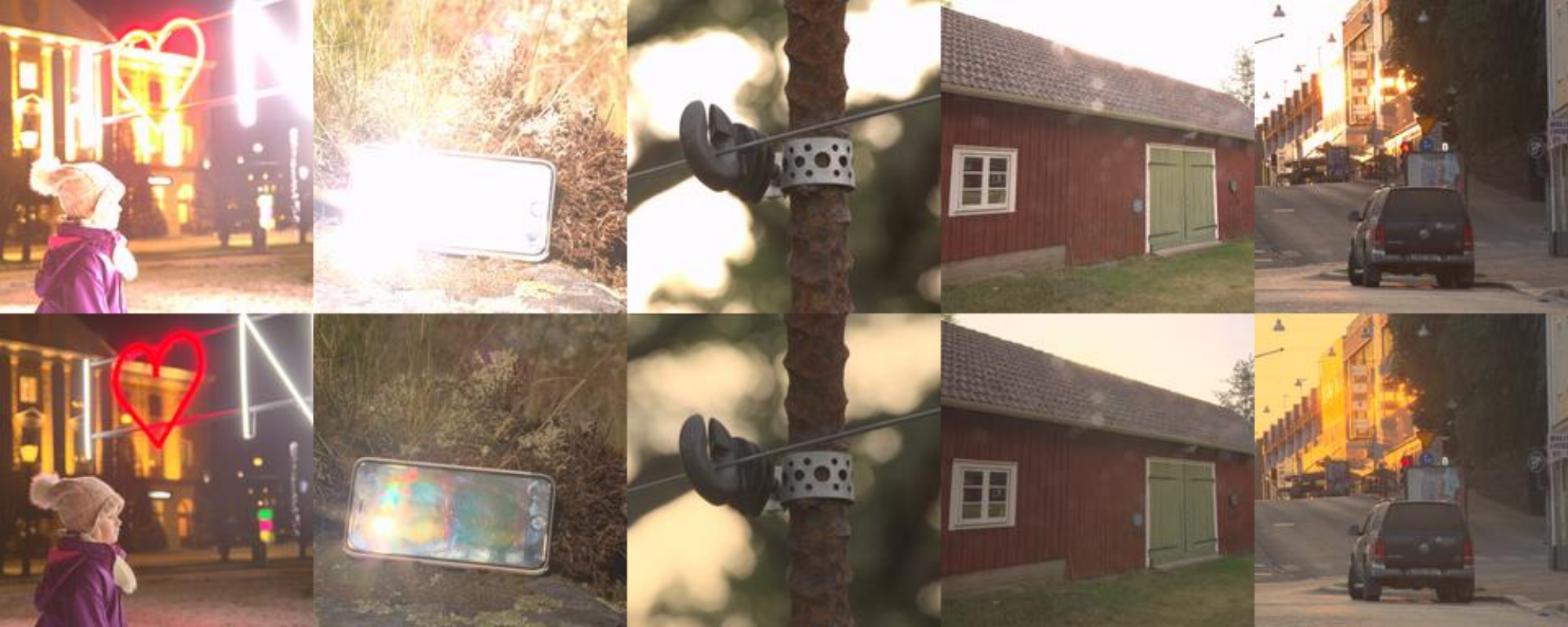}
    \caption{\textbf{Bright-input scenes where our method outperforms most baselines.}
Input LDR thumbnails (top row) and tone-mapped ground-truth HDR images (bottom row)
for the 10 brightly exposed test scenes (mean LDR luminance $\geq 0.40$). These examples
show that over-exposure can reduce apparent edge density in the input LDR, even when
the underlying HDR scene contains substantial structure. In such bright-input,
moderate-to-low-edge-density cases, our method outperforms most baselines.}
    \label{fig:bright-wins}
\end{figure}

The latter case is especially important for single-shot HDR reconstruction. In \cref{fig:bright-wins}, we show additional over-exposed input LDRs for scenes where our method outperforms the baselines. These examples indicate that over-exposure can reduce apparent edge density in the input image, even when the underlying HDR scene contains substantial structure. Compared to the failure cases in \cref{fig:worst-results}, these scenes contain fewer high-frequency textures such as foliage, which helps explain why our method performs better in this regime.

The edge-density analysis also highlights the 
failure regime, i.e., appropriately exposed scenes with high-edge-density. Since these scenes
require preserving fine spatial detail rather than recovering missing dynamic
range, it is natural to investigate the effect of the SVD VAE, which compresses
images into a lower-dimensional latent representation. This motivates our VAE
round-trip analysis in \cref{supp:vae-roundtrip}, where we show that the
latent bottleneck can attenuate high-frequency texture.

\paragraph{Qualitative failure cases:}
Fig.~\ref{fig:worst-results} shows the five scenes on which our method trails
existing methods by the largest PSNR margin. Most of them are dense outdoor or
textured-foliage scenes. While our reconstructions visibly lose a small
amount of fine high-frequency detail relative to the strongest baselines,
they remain visually comparable overall. This is consistent with our user
study (\cref{tab:user-study}): even across scenes of this type, our
method is preferred over the baselines $72\%$ of the time on average. The scenes where our method is least preferred in the user study (\cref{fig:userstudy-worst}) show qualitative similarities to the well-exposed cases where it underperforms the baselines in PSNR (\cref{fig:worst-results}). 
Both sets of failure cases are well-exposed scenes with dense high-frequency detail, such as foliage and tiled patterns.

\begin{table}
\centering
\small
\begin{tabular}{l|ccc}
\toprule
\textbf{Scene set} & \textbf{$n$} & \textbf{$\rho_{\text{VAE}} \uparrow$} & \textbf{$\rho_{\text{full}} \uparrow$} \\
\midrule
Best-10  (smooth / dark scenes)        & 10 & 0.89 & 0.72 \\
Worst-10 (textured outdoor scenes)     & 10 & 0.69 & 0.45 \\
\bottomrule
\end{tabular}
\caption{\textbf{VAE-induced high-frequency energy loss.}
$\rho_{\text{VAE}}$ is the ratio between the high-frequency energy (Laplacian std of luminance) of a GT image that has been PU-21-encoded, encoded--decoded through our SVD VAE, and compared back in PU-21 space, and that of the original PU-21-encoded GT. $\rho_{\text{full}}$ is the same ratio computed on our predicted HDR (also PU-21 encoded). Encoding inputs in PU-21 places the VAE in its trained distribution range and isolates its compression bottleneck from input-distribution artifacts. We report the mean over the 10 scenes where our method most outperforms baselines (Best-10) and the 10 where it most underperforms (Worst-10). The VAE alone discards $31\%$ of input HF energy on textured scenes versus only $11\%$ on smooth scenes, accounting for $\approx 75\%$ of the cross-scene gap in our pipeline's HF preservation.}
\label{tab:vae-roundtrip}
\end{table}

\subsection{Quantifying VAE-induced high-frequency attenuation}
\label{supp:vae-roundtrip}
We analyze the contribution of the SVD VAE to the loss of fine detail observed
on well-exposed, high-edge-density inputs. For each ground-truth HDR image in
the SI-HDR test set, we clip intensities to the $99$th percentile, rescale the
peak to $4000$\,cd/m$^2$, and apply PU-21 encoding to obtain a perceptually
uniform image in the range $[0,1]$. We then encode and decode this image
with the SVD VAE encoder and decoder, respectively. Using PU-21 avoids confounding VAE compression artifacts with the distribution mismatch between the training domain of the VAE and HDR images~\citet{wu2026x2hdr}.

To measure detail preservation, we compute the luminance of each image
and apply a $3{\times}3$ Laplacian filter; the standard deviation of this
Laplacian filtered image serves as our high-frequency energy measure. We report $\rho_{\text{VAE}}$, the ratio of this measure after VAE reconstruction to that of the original image. $\rho_{\text{VAE}}<1$ indicates high-frequency detail lost
solely by the VAE. For comparison, we compute the same ratio for our full HDR
pipeline, $\rho_{\text{full}}$, after PU-21 encoding the predicted HDR output.

We average these ratios over two subsets: the $10$ scenes where our method
outperforms baselines on PSNR, and the $10$ scenes where it underperforms the
baselines. Results are shown in \cref{tab:vae-roundtrip}. On textured
scenes, the VAE preserves only $69\%$ of the high-frequency measure, which
drops to $45\%$ after the full pipeline. On smoother scenes (due to
over/under-exposure), the VAE preserves $89\%$, dropping to $72\%$ after the
full pipeline. The cross-scene gap in $\rho_{\text{VAE}}$ accounts for roughly
$75\%$ of the gap in $\rho_{\text{full}}$, indicating that the SVD VAE is a
major contributor to the high-frequency deficit on textured scenes.

This helps explain why non-generative methods such as SingleHDR and HDRCNN, or
methods such as BracketDiff that do not rely on a latent VAE bottleneck,
can potentially outperform our method on high-texture, well-exposed inputs.
Importantly, this failure mode is tied to the VAE compression stage rather than
to the exposure-bracket generation formulation itself. Moreover, these failure
cases are largely well-exposed scenes that already preserve most visible
content, and therefore benefit less from dynamic range expansion.

\section{Comparing with Flux-Based X2HDR}
\label{app:x2hdr-flux}
The original X2HDR-Flux model fine-tunes FLUX.1-dev, which uses a 12B-parameter rectified-flow transformer as its denoising network. In contrast, our method fine-tunes the 1.5B-parameter SVD diffusion UNet. We report a comparison of our method with X2HDR-Flux in \cref{tab:hdr-x2hdr}. The gap between X2HDR-SD2.1 and X2HDR-Flux indicates that backbone capacity is an important factor for HDR reconstruction. Nevertheless, our method outperforms both X2HDR variants on $Q^{\ast}$-PSNR and $Q^{\ast}$-MAE, while remaining comparable on Q-JOD. This suggests that explicitly generating an exposure bracket followed by learned fusion provides a strong reconstruction formulation, even when using a substantially smaller diffusion backbone. Scaling our method to video diffusion models with larger backbones would thus be an interesting step in further improving performance. 
\begin{table}
\centering
\small
\setlength{\tabcolsep}{4pt}
\begin{tabular}{l|ccccc}
\toprule
\textbf{Method} & \textbf{$Q^{\ast}$-PSNR $\uparrow$} & \textbf{$Q^{\ast}$-SSIM $\uparrow$} & \textbf{$Q^{\ast}$-MAE $\downarrow$} & \textbf{$Q^{\ast}$-LPIPS $\downarrow$} & \textbf{Q-JOD $\uparrow$} \\
\midrule
Ours          & $\mathbf{22.36 \pm 3.48}$  & $0.664 \pm 0.156$          & $\mathbf{0.060 \pm 0.027}$ & $0.289 \pm 0.111$          & $7.51 \pm 0.99$ \\
X2HDR-SD2.1 & $21.24 \pm 3.32$ & $0.585 \pm 0.150$ & $0.068 \pm 0.030$ & $0.335 \pm 0.120$ & $7.17 \pm 0.97$ \\
X2HDR-Flux  & $21.37 \pm 3.10$           & $\mathbf{0.726 \pm 0.108}$ & $0.067 \pm 0.025$          & $\mathbf{0.232 \pm 0.106}$ & $\mathbf{7.59 \pm 1.24}$ \\
\bottomrule
\end{tabular}
\caption{\textbf{Comparison with different X2HDR backbones.} We compare against both our controlled X2HDR-SD2.1 implementation and the original Flux-based X2HDR model. Our method achieves the best $Q^{\ast}$-PSNR and $Q^{\ast}$-MAE, while X2HDR-Flux performs best on $Q^{\ast}$-SSIM, Q-JOD and $Q^{\ast}$-LPIPS.}
\label{tab:hdr-x2hdr}
\end{table}
\newpage
\section*{NeurIPS Paper Checklist}
\begin{enumerate}
\item {\bf Claims}
    \item[] Question: Do the main claims made in the abstract and introduction accurately reflect the paper's contributions and scope?
    \item[] Answer:. \answerYes 
    \item[] Justification:  We demonstrate evidence of the claims in the abstract in the Experiments and Results section. 
    \item[] Guidelines:
    \begin{itemize}
        \item The answer NA means that the abstract and introduction do not include the claims made in the paper.
        \item The abstract and/or introduction should clearly state the claims made, including the contributions made in the paper and important assumptions and limitations. A No or NA answer to this question will not be perceived well by the reviewers. 
        \item The claims made should match theoretical and experimental results, and reflect how much the results can be expected to generalize to other settings. 
        \item It is fine to include aspirational goals as motivation as long as it is clear that these goals are not attained by the paper. 
    \end{itemize}

\item {\bf Limitations}
    \item[] Question: Does the paper discuss the limitations of the work performed by the authors?
    \item[] Answer: \answerYes{} 
    \item[] Justification: In the limitations and future work section, we discuss the limitations of our method. We also visualize some failure cases in Figure 5. 
    
    \item[] Guidelines:
    \begin{itemize}
        \item The answer NA means that the paper has no limitation while the answer No means that the paper has limitations, but those are not discussed in the paper. 
        \item The authors are encouraged to create a separate "Limitations" section in their paper.
        \item The paper should point out any strong assumptions and how robust the results are to violations of these assumptions (e.g., independence assumptions, noiseless settings, model well-specification, asymptotic approximations only holding locally). The authors should reflect on how these assumptions might be violated in practice and what the implications would be.
        \item The authors should reflect on the scope of the claims made, e.g., if the approach was only tested on a few datasets or with a few runs. In general, empirical results often depend on implicit assumptions, which should be articulated.
        \item The authors should reflect on the factors that influence the performance of the approach. For example, a facial recognition algorithm may perform poorly when image resolution is low or images are taken in low lighting. Or a speech-to-text system might not be used reliably to provide closed captions for online lectures because it fails to handle technical jargon.
        \item The authors should discuss the computational efficiency of the proposed algorithms and how they scale with dataset size.
        \item If applicable, the authors should discuss possible limitations of their approach to address problems of privacy and fairness.
        \item While the authors might fear that complete honesty about limitations might be used by reviewers as grounds for rejection, a worse outcome might be that reviewers discover limitations that aren't acknowledged in the paper. The authors should use their best judgment and recognize that individual actions in favor of transparency play an important role in developing norms that preserve the integrity of the community. Reviewers will be specifically instructed to not penalize honesty concerning limitations.
    \end{itemize}

\item {\bf Theory assumptions and proofs}
    \item[] Question: For each theoretical result, does the paper provide the full set of assumptions and a complete (and correct) proof?
    \item[] Answer: \answerNA{} 
    \item[] Guidelines:
    \begin{itemize}
        \item The answer NA means that the paper does not include theoretical results. 
        \item All the theorems, formulas, and proofs in the paper should be numbered and cross-referenced.
        \item All assumptions should be clearly stated or referenced in the statement of any theorems.
        \item The proofs can either appear in the main paper or the supplemental material, but if they appear in the supplemental material, the authors are encouraged to provide a short proof sketch to provide intuition. 
        \item Inversely, any informal proof provided in the core of the paper should be complemented by formal proofs provided in appendix or supplemental material.
        \item Theorems and Lemmas that the proof relies upon should be properly referenced. 
    \end{itemize}

    \item {\bf Experimental result reproducibility}
    \item[] Question: Does the paper fully disclose all the information needed to reproduce the main experimental results of the paper to the extent that it affects the main claims and/or conclusions of the paper (regardless of whether the code and data are provided or not)?
    \item[] Answer: \answerYes{} 
     \item[] Justification: We have mentioned the major hyperparameters in the Implementation details section of the paper.
    \item[] Guidelines:
    \begin{itemize}
        \item The answer NA means that the paper does not include experiments.
        \item If the paper includes experiments, a No answer to this question will not be perceived well by the reviewers: Making the paper reproducible is important, regardless of whether the code and data are provided or not.
        \item If the contribution is a dataset and/or model, the authors should describe the steps taken to make their results reproducible or verifiable. 
        \item Depending on the contribution, reproducibility can be accomplished in various ways. For example, if the contribution is a novel architecture, describing the architecture fully might suffice, or if the contribution is a specific model and empirical evaluation, it may be necessary to either make it possible for others to replicate the model with the same dataset, or provide access to the model. In general. releasing code and data is often one good way to accomplish this, but reproducibility can also be provided via detailed instructions for how to replicate the results, access to a hosted model (e.g., in the case of a large language model), releasing of a model checkpoint, or other means that are appropriate to the research performed.
        \item While NeurIPS does not require releasing code, the conference does require all submissions to provide some reasonable avenue for reproducibility, which may depend on the nature of the contribution. For example
        \begin{enumerate}
            \item If the contribution is primarily a new algorithm, the paper should make it clear how to reproduce that algorithm.
            \item If the contribution is primarily a new model architecture, the paper should describe the architecture clearly and fully.
            \item If the contribution is a new model (e.g., a large language model), then there should either be a way to access this model for reproducing the results or a way to reproduce the model (e.g., with an open-source dataset or instructions for how to construct the dataset).
            \item We recognize that reproducibility may be tricky in some cases, in which case authors are welcome to describe the particular way they provide for reproducibility. In the case of closed-source models, it may be that access to the model is limited in some way (e.g., to registered users), but it should be possible for other researchers to have some path to reproducing or verifying the results.
        \end{enumerate}
    \end{itemize}

\item {\bf Open access to data and code}
    \item[] Question: Does the paper provide open access to the data and code, with sufficient instructions to faithfully reproduce the main experimental results, as described in the supplemental material?
    \item[] Answer: \answerYes{} 
    \item[] Justification: We will release the code after acceptance. 
    \item[] Guidelines:
    \begin{itemize}
        \item The answer NA means that paper does not include experiments requiring code.
        \item Please see the NeurIPS code and data submission guidelines (\url{https://nips.cc/public/guides/CodeSubmissionPolicy}) for more details.
        \item While we encourage the release of code and data, we understand that this might not be possible, so “No” is an acceptable answer. Papers cannot be rejected simply for not including code, unless this is central to the contribution (e.g., for a new open-source benchmark).
        \item The instructions should contain the exact command and environment needed to run to reproduce the results. See the NeurIPS code and data submission guidelines (\url{https://nips.cc/public/guides/CodeSubmissionPolicy}) for more details.
        \item The authors should provide instructions on data access and preparation, including how to access the raw data, preprocessed data, intermediate data, and generated data, etc.
        \item The authors should provide scripts to reproduce all experimental results for the new proposed method and baselines. If only a subset of experiments are reproducible, they should state which ones are omitted from the script and why.
        \item At submission time, to preserve anonymity, the authors should release anonymized versions (if applicable).
        \item Providing as much information as possible in supplemental material (appended to the paper) is recommended, but including URLs to data and code is permitted.
    \end{itemize}

\item {\bf Experimental setting/details}
    \item[] Question: Does the paper specify all the training and test details (e.g., data splits, hyperparameters, how they were chosen, type of optimizer, etc.) necessary to understand the results?
    \item[] Answer: \answerYes{} 
    \item[] Justification: We specify implementation details in the paper; additional details will be included in the supplement. 
    \item[] Guidelines:
    \begin{itemize}
        \item The answer NA means that the paper does not include experiments.
        \item The experimental setting should be presented in the core of the paper to a level of detail that is necessary to appreciate the results and make sense of them.
        \item The full details can be provided either with the code, in appendix, or as supplemental material.
    \end{itemize}

\item {\bf Experiment statistical significance}
    \item[] Question: Does the paper report error bars suitably and correctly defined or other appropriate information about the statistical significance of the experiments?
    \item[] Answer: \answerYes{} 
    \item[] Justification: We report standard deviations for the performance metrics over the test set in the main results table.
    \item[] Guidelines:
    \begin{itemize}
        \item The answer NA means that the paper does not include experiments.
        \item The authors should answer "Yes" if the results are accompanied by error bars, confidence intervals, or statistical significance tests, at least for the experiments that support the main claims of the paper.
        \item The factors of variability that the error bars are capturing should be clearly stated (for example, train/test split, initialization, random drawing of some parameter, or overall run with given experimental conditions).
        \item The method for calculating the error bars should be explained (closed form formula, call to a library function, bootstrap, etc.)
        \item The assumptions made should be given (e.g., Normally distributed errors).
        \item It should be clear whether the error bar is the standard deviation or the standard error of the mean.
        \item It is OK to report 1-sigma error bars, but one should state it. The authors should preferably report a 2-sigma error bar than state that they have a 96\% CI, if the hypothesis of Normality of errors is not verified.
        \item For asymmetric distributions, the authors should be careful not to show in tables or figures symmetric error bars that would yield results that are out of range (e.g. negative error rates).
        \item If error bars are reported in tables or plots, The authors should explain in the text how they were calculated and reference the corresponding figures or tables in the text.
    \end{itemize}

\item {\bf Experiments compute resources}
    \item[] Question: For each experiment, does the paper provide sufficient information on the computer resources (type of compute workers, memory, time of execution) needed to reproduce the experiments?
    \item[] Answer: \answerYes{} 
    \item[] Justification: We mention the GPU used, memory usage, and optimization time for our algorithm. 
    \item[] Guidelines:
    \begin{itemize}
        \item The answer NA means that the paper does not include experiments.
        \item The paper should indicate the type of compute workers CPU or GPU, internal cluster, or cloud provider, including relevant memory and storage.
        \item The paper should provide the amount of compute required for each of the individual experimental runs as well as estimate the total compute. 
        \item The paper should disclose whether the full research project required more compute than the experiments reported in the paper (e.g., preliminary or failed experiments that didn't make it into the paper). 
    \end{itemize}
    
\item {\bf Code of ethics}
    \item[] Question: Does the research conducted in the paper conform, in every respect, with the NeurIPS Code of Ethics \url{https://neurips.cc/public/EthicsGuidelines}?
    \item[] Answer: \answerYes{} 
    \item[] Justification: We conform to the ethics guidelines. 
    \item[] Guidelines:
    \begin{itemize}
        \item The answer NA means that the authors have not reviewed the NeurIPS Code of Ethics.
        \item If the authors answer No, they should explain the special circumstances that require a deviation from the Code of Ethics.
        \item The authors should make sure to preserve anonymity (e.g., if there is a special consideration due to laws or regulations in their jurisdiction).
    \end{itemize}

\item {\bf Broader impacts}
    \item[] Question: Does the paper discuss both potential positive societal impacts and negative societal impacts of the work performed?
    \item[] Answer: \answerNA{} 
    \item[] Justification: Our work has no particularly negative societal impacts. 
    \item[] Guidelines:
    \begin{itemize}
        \item The answer NA means that there is no societal impact of the work performed.
        \item If the authors answer NA or No, they should explain why their work has no societal impact or why the paper does not address societal impact.
        \item Examples of negative societal impacts include potential malicious or unintended uses (e.g., disinformation, generating fake profiles, surveillance), fairness considerations (e.g., deployment of technologies that could make decisions that unfairly impact specific groups), privacy considerations, and security considerations.
        \item The conference expects that many papers will be foundational research and not tied to particular applications, let alone deployments. However, if there is a direct path to any negative applications, the authors should point it out. For example, it is legitimate to point out that an improvement in the quality of generative models could be used to generate deepfakes for disinformation. On the other hand, it is not needed to point out that a generic algorithm for optimizing neural networks could enable people to train models that generate Deepfakes faster.
        \item The authors should consider possible harms that could arise when the technology is being used as intended and functioning correctly, harms that could arise when the technology is being used as intended but gives incorrect results, and harms following from (intentional or unintentional) misuse of the technology.
        \item If there are negative societal impacts, the authors could also discuss possible mitigation strategies (e.g., gated release of models, providing defenses in addition to attacks, mechanisms for monitoring misuse, mechanisms to monitor how a system learns from feedback over time, improving the efficiency and accessibility of ML).
    \end{itemize}
    
\item {\bf Safeguards}
    \item[] Question: Does the paper describe safeguards that have been put in place for responsible release of data or models that have a high risk for misuse (e.g., pretrained language models, image generators, or scraped datasets)?
    \item[] Answer: \answerNA{} 
    \item[] Justification: Our work poses no such risks. 
    \item[] Guidelines:
    \begin{itemize}
        \item The answer NA means that the paper poses no such risks.
        \item Released models that have a high risk for misuse or dual-use should be released with necessary safeguards to allow for controlled use of the model, for example by requiring that users adhere to usage guidelines or restrictions to access the model or implementing safety filters. 
        \item Datasets that have been scraped from the Internet could pose safety risks. The authors should describe how they avoided releasing unsafe images.
        \item We recognize that providing effective safeguards is challenging, and many papers do not require this, but we encourage authors to take this into account and make a best faith effort.
    \end{itemize}

\item {\bf Licenses for existing assets}
    \item[] Question: Are the creators or original owners of assets (e.g., code, data, models), used in the paper, properly credited and are the license and terms of use explicitly mentioned and properly respected?
    \item[] Answer:\answerYes{} 
    \item[] Justification: We have cited the datasets and models we used for comparison. These datasets have been widely used in published papers on this topic. 
    \item[] Guidelines:
    \begin{itemize}
        \item The answer NA means that the paper does not use existing assets.
        \item The authors should cite the original paper that produced the code package or dataset.
        \item The authors should state which version of the asset is used and, if possible, include a URL.
        \item The name of the license (e.g., CC-BY 4.0) should be included for each asset.
        \item For scraped data from a particular source (e.g., website), the copyright and terms of service of that source should be provided.
        \item If assets are released, the license, copyright information, and terms of use in the package should be provided. For popular datasets, \url{paperswithcode.com/datasets} has curated licenses for some datasets. Their licensing guide can help determine the license of a dataset.
        \item For existing datasets that are re-packaged, both the original license and the license of the derived asset (if it has changed) should be provided.
        \item If this information is not available online, the authors are encouraged to reach out to the asset's creators.
    \end{itemize}

\item {\bf New assets}
    \item[] Question: Are new assets introduced in the paper well documented and is the documentation provided alongside the assets?
    \item[] Answer: \answerYes{} 
    \item[] Justification: We will release our fine-tuned model with the code release of the paper.
    \item[] Guidelines:
    \begin{itemize}
        \item The answer NA means that the paper does not release new assets.
        \item Researchers should communicate the details of the dataset/code/model as part of their submissions via structured templates. This includes details about training, license, limitations, etc. 
        \item The paper should discuss whether and how consent was obtained from people whose asset is used.
        \item At submission time, remember to anonymize your assets (if applicable). You can either create an anonymized URL or include an anonymized zip file.
    \end{itemize}

\item {\bf Crowdsourcing and research with human subjects}
    \item[] Question: For crowdsourcing experiments and research with human subjects, does the paper include the full text of instructions given to participants and screenshots, if applicable, as well as details about compensation (if any)? 
    \item[] Answer: \answerYes{} 
    \item[] Justification: Yes we will include these in the supplement 
    \item[] Guidelines:
    \begin{itemize}
        \item The answer NA means that the paper does not involve crowdsourcing nor research with human subjects.
        \item Including this information in the supplemental material is fine, but if the main contribution of the paper involves human subjects, then as much detail as possible should be included in the main paper. 
        \item According to the NeurIPS Code of Ethics, workers involved in data collection, curation, or other labor should be paid at least the minimum wage in the country of the data collector. 
    \end{itemize}

\item {\bf Institutional review board (IRB) approvals or equivalent for research with human subjects}
    \item[] Question: Does the paper describe potential risks incurred by study participants, whether such risks were disclosed to the subjects, and whether Institutional Review Board (IRB) approvals (or an equivalent approval/review based on the requirements of your country or institution) were obtained?
    \item[] Answer: \answerYes{} 
    \item[] Justification: Yes, we obtained an IRB exemption for our study from our institute. 
    \item[] Guidelines:
    \begin{itemize}
        \item The answer NA means that the paper does not involve crowdsourcing nor research with human subjects.
        \item Depending on the country in which research is conducted, IRB approval (or equivalent) may be required for any human subjects research. If you obtained IRB approval, you should clearly state this in the paper. 
        \item We recognize that the procedures for this may vary significantly between institutions and locations, and we expect authors to adhere to the NeurIPS Code of Ethics and the guidelines for their institution. 
        \item For initial submissions, do not include any information that would break anonymity (if applicable), such as the institution conducting the review.
    \end{itemize}

\item {\bf Declaration of LLM usage}
    \item[] Question: Does the paper describe the usage of LLMs if it is an important, original, or non-standard component of the core methods in this research? Note that if the LLM is used only for writing, editing, or formatting purposes and does not impact the core methodology, scientific rigor, or originality of the research, a declaration is not required.
    \item[] Answer: \answerNA{} 
    \item[] Justification: This paper is not related to LLM research or applications. 
    \item[] Guidelines:
    \begin{itemize}
        \item The answer NA means that the core method development in this research does not involve LLMs as any important, original, or non-standard components.
        \item Please refer to our LLM policy (\url{https://neurips.cc/Conferences/2025/LLM}) for what should or should not be described.
    \end{itemize}

\end{enumerate}

\end{document}